\documentclass[preprint,12pt]{elsarticle}
\usepackage[pdfborder={0 0 0}]{hyperref}
\usepackage{amsthm,amsmath,amsfonts,latexsym,amssymb}
\usepackage{graphics,fancyhdr,graphicx,subfigure}
\usepackage[ruled,linesnumbered]{algorithm2e}
\usepackage[margin=1in]{geometry}
\usepackage{multirow,booktabs}
\usepackage[table]{xcolor}
\usepackage{listings}
\usepackage{tabularx}
\usepackage{placeins}
\usepackage{comment}
\usepackage{lineno}
\usepackage{lscape}
\usepackage{bm}
\usepackage{hyperref}
\usepackage{mathrsfs}
\usepackage{arydshln}
\usepackage{appendix}
\definecolor{custom-blue}{RGB}{3,69,173}
\hypersetup{
    colorlinks = true,
    urlcolor   = blue,
    citecolor  = custom-blue,
}

\biboptions{numbers,sort&compress}
\definecolor{listinggray}{gray}{0.9}
\definecolor{lbcolor}{rgb}{0.9,0.9,0.9}
\definecolor{Darkgreen}{RGB}{0,100,0}

\begin{document}
\abovedisplayskip=6.0pt
\belowdisplayskip=6.0pt
\begin{frontmatter}

\title{On the influence of over-parameterization in manifold based surrogates and deep neural operators}

\author[1]{Katiana Kontolati}
\ead{kontolati@jhu.edu}
\author[2]{Somdatta Goswami}
\ead{somdatta_goswami@brown.edu}
\author[1]{Michael D. Shields}
\ead{michael.shields@jhu.edu}
\author[2,3]{George Em Karniadakis\corref{cor1}}
\ead{george_karniadakis@brown.edu}

\address[1]{Department of Civil and Systems Engineering, Johns Hopkins University}
\address[2]{Division of Applied Mathematics, Brown University}
\address[3]{School of Engineering, Brown University}
\cortext[cor1]{Corresponding author.}

\begin{abstract}
\noindent
Constructing accurate and generalizable approximators (surrogate models) for complex physico-chemical processes exhibiting highly non-smooth dynamics is challenging. The main question is what type of surrogate models we should construct and should these models be under-parameterized or over-parameterized. In this work, we propose new developments and perform comparisons for two promising approaches: manifold-based polynomial chaos expansion (m-PCE) and the deep neural operator (DeepONet), and we examine the effect of over-parameterization on generalization. While m-PCE enables the construction of a mapping by first identifying low-dimensional embeddings of the input functions, parameters, and quantities of interest (QoIs), a neural operator learns the nonlinear mapping via the use of deep neural networks. We demonstrate the performance of these methods in terms of generalization accuracy by solving the 2D time-dependent Brusselator reaction-diffusion system with uncertainty sources, modeling an autocatalytic chemical reaction between two species. We first propose an extension of the m-PCE by constructing a mapping between latent spaces formed by two separate embeddings of the input functions and the output QoIs. To further enhance the accuracy of the DeepONet, we introduce weight self-adaptivity in the loss function. We demonstrate that the performance of m-PCE and DeepONet is comparable for cases of relatively smooth input-output mappings. However, when highly non-smooth dynamics is considered, DeepONet shows higher approximation accuracy. We also find that for m-PCE, modest over-parameterization leads to better generalization, both within and outside of distribution, whereas aggressive over-parameterization leads to over-fitting. In contrast, an even highly over-parameterized DeepONet leads to better generalization for both smooth and non-smooth dynamics. Furthermore, we  compare the performance of the above models with another recently proposed operator learning model, the Fourier Neural Operator, and show that its over-parameterization also leads to better generalization. Taken together, our studies show that m-PCE can provide very good accuracy at very low training cost, whereas a highly over-parameterized DeepONet can provide better accuracy and robustness to noise but at higher training cost. In both methods, the inference cost is negligible.
\end{abstract}


\end{frontmatter}


\section{Introduction}
\label{sec:intro}

Surrogate models serve as efficient approximations for expensive high-fidelity simulations, with the goal of providing significant computational savings while maintaining solution accuracy. Data-driven surrogate modeling approaches like Gaussian process regression (GPR)\cite{chen2015uncertainty, tripathy2016gaussian, raissi2018numerical}, polynomial chaos expansion (PCE) \cite{ghanem1990polynomial, xiu2002wiener, oladyshkin2012data, zheng2015adaptive}, response surface methods\cite{goswami2016reliability,goswami2013adaptive}, manifold-based approaches \cite{giovanis2020data, kontolati2021manifold}, and deep neural networks (DNNs) \cite{di2021deeponet, olivier2021bayesian} are gaining popularity as a means of speeding up the engineering design/analysis process. For theoretical validity, training and test data sets for surrogate model construction must have the same support. However, in many practical uses, test data are drawn from outside the compact support of the training data -- resulting in the use of the surrogate model for prediction on cases for which it has not been trained (i.e. extrapolation rather than interpolation). This may occur, for example, when test data are drawn from a different probability distribution than the training data (so-called out-of-distribution, OOD, data) or when input are degraded by added noise. The ability to construct surrogate models that generalize beyond the immediate support of their training data and into noisy data regimes is particularly challenging and is expressly connected to the parameterization of the surrogate. Resolving this challenge is of broad interest to the physics-based modeling community. 

GPR and PCE-based surrogates are particularly favorable because they involve relatively few hyperparameters and thus are easy to implement. However, they fail to handle physical domains with complex geometry, unseen noisy input data, and suffer from the curse-of-dimensionality. Manifold-based approaches, on the other hand, can handle high-dimensional inputs \cite{lataniotis2020extending}. A recent study on manifold-based PCE (\textit{m-PCE}) \cite{kontolati2022survey} compares several manifold learning methods and provides insights into their effectiveness for the construction of accurate and generalizable surrogates for physics-based models. The key benefits of these manifold-based methods, and \textit{m-PCE} in particular, are their predictive accuracy, robustness to noisy input data, limited number of tunable hyperparameters, and computational efficiency during training. Alternatively, DNNs have proven to be  effective surrogate models across a broad spectrum of applications \cite{raissi2019physics,tartakovsky2020physics,chen2020generative,jiang2021deep,lu2021learning,goswami2022physics}. The pivotal work of \cite{krizhevsky2012imagenet} kicked off an intense period of research to train larger networks with more hidden units in the search of higher test performance. Despite the fact that the high complexity of such over-parameterized models allows for faultless data interpolation, they also frequently achieve low generalization error \cite{lanthaler2021error}.

Conventional thinking in machine learning holds that using models with increasing capacity will lead to overfitting to the training data. As a result, the capacity of the model is generally controlled either by limiting the size of the model (number of parameters) or by introducing additional explicit regularization terms to avoid overfitting. However, for DNNs, expanding the model size only improves the generalization error, even when the networks are trained without any regularization term \cite{neyshabur2018role, du2018power,poggio2020theoretical,dar2021farewell}. Over-parameterization refers to the case where the number of trainable parameters of a model/network is larger than the number of training observations. While over-parameterized DNNs are the state-of-the-art in machine learning, their robustness and ability to generalize even to noisy test data is what makes this architecture stand out. Optimization is usually easier in over-parameterized models \cite{poggio2020theoretical}. The idea is that over-parameterization changes the loss function, resulting in a large number of locally optimum solutions, making it easier for local search algorithms to locate an near-optimal solution. However, it has been shown that the computational cost of training an over-parameterized model grows at least as a fourth-order polynomial with respect to performance, $\it{i.e.}$ Computation = $\mathcal{O}$(Performance$^4$), where performance refers to an error measure, for example, root mean squared error, and this may even be an underestimation \cite{thompson2020computational}. On the other side of the spectrum, under-parameterized models are characterized by limited expressivity, but are often preferred due to their simplicity, interpretability, computational efficiency, and generalizability for smooth functions. In general, when small training sets are available, adding more features yields improvements in the training process \cite{belkin2019reconciling}. 

Nevertheless, model complexity and efficiency cannot be measured solely by the total number of trainable parameters since the expressivity of these parameters should be accounted for as well. For example, in neural network models increasing the depth of the network by adding new layers might lead to higher expressivity than increasing the size of existing layers even if the total number of parameters is the same. In addition, the ratio between the number of data features and number of data should also be considered. Therefore it is imperative to understand how all the aforementioned affect the robustness of the surrogate model \cite{bubeck2021universal}. 

In this work, we present a systematic study comparing under- and over-parameterized manifold-based PCE methods with over-parameterized DNNs. Through this study, we have tried to answer basic questions about how parameterization of these models relates to their ability to generalize to OOD and noisy data. To do so, we specifically investigated the influence of various factors such as model complexity, label noise and dataset size in the construction of surrogate models. From the corpus of the available models, we study under- and over-parameterization performance experiments using \textit{m-PCE} with kernel-PCA and over-parameterization experiments on two state-of-the-art neural operators, the Deep Operator Network (\textit{DeepONet}) \cite{lu2021learning} and the Fourier Neural Operator (\textit{FNO}) \cite{li2020fourier}. The comparative study includes a recent extension of the DeepONet, namely POD-DeepONet \cite{lu2021comprehensive}, while for FNO we have employed the version from \url{https://github.com/zongyi-li/fourier_neural_operator}. Additionally, we introduce some enhancements for both manifold-based surrogates and DeepONet to allow fast training and convergence and better generalization for problems with highly nonlinear and nonsmooth solutions. The new developments proposed in the current work are the following:
\begin{itemize}
    \item We perform manifold-learning for both the model inputs and model outputs for construction of both over- and under-parameterized \textit{m-PCE} surrogates.
    \item We introduce fully-trainable weight parameters for the DeepONet, referred to as \textit{DeepONet with self-adaptivity}, to handle highly nonlinear and non-smooth data.
\end{itemize}

We have considered the Brusselator diffusion-reaction dynamical system to investigate the relative performance of the chosen surrogate models. All codes will become available on Github, \url{https://github.com/katiana22/surrogate-overparameterization}, upon publication of our paper.  Construction of PCE surrogates is performed using the open-source UQpy (Uncertainty Quantification with Python) \cite{olivier2020uqpy} package. The paper is organized as follows.  In Section \ref{sec:methods}, we describe the DeepONet and FNO architectures, present the manifold-based PCE models and introduce the proposed enhancements. In Section \ref{sec:bruss}, we present the Brusselator system and the data generation process. In Section \ref{sec:results}, we compare the performance of the studied models for the Brusselator diffusion reaction system. Finally, we summarize our observations and provide concluding remarks in Section \ref{sec:summary_and_discussions}.

\section{Machine Learned Approximation Methods}
\label{sec:methods}

Consider an analytical or a computational model, $\mathcal{M}(\mathbf{x})$, which simulates a physical process and represents a mapping between a vector of input random variables, $\mathbf{x}(\zeta) \in \mathbb{R}^{D_{\text{in}}}$, and corresponding output quantities of interest (QoIs), $\mathbf{y}(\zeta) \in \mathbb{R}^{D_{\text{out}}}$ where $\zeta$ represent spatio-temporal coordinates. 
That is, model $\mathcal{M}$ performs the mapping $\mathcal{M}:{\mathbf{x} \in \mathbb{R}^{D_{\text{in}}}} \rightarrow \mathbf{y} \in {\mathbb{R}^{D_{\text{out}}}}$ where the dimensionality of both the inputs and the outputs, $D_{\text{in}}, D_{\text{out}}$ is high, $\textit{e.g.}$ $\mathcal{O}(10^{2-4})$. Here, the high-dimensional inputs may represent random fields and/or processes, such as spatially or temporally varying coefficients, and the corresponding QoIs represent physical quantities, which can also vary in both space and time. Our objective is to approximate the mapping, $\mathcal{M}$ from a training dataset of $N$ input-output pairs $(\mathbf{X},\mathbf{Y})$ where $\mathbf{X} = \{\mathbf{x}_{1},..,\mathbf{x}_{N}\}$ and $\mathbf{Y} = \{\mathbf{y}_{1},..,\mathbf{y}_{N}\}$, and achieve the lowest possible predictive error on a test dataset. In this section, we outline two classes of approaches to learn this approximation: operator learning and manifold learning based surrogates. We then systematically test these approximations for generalization error on out-of-distribution (OOD) and noisy input data for a set of nonlinear PDEs -- the Brusselator reaction-diffusion equations -- in subsequent sections.

\subsection{Operator learning}
\label{sec:operator-learning}

Neural operators learn nonlinear mappings between infinite dimensional function spaces on bounded domains, providing a unique simulation framework for real-time prediction of multi-dimensional complex dynamics. Once trained, such models are discretization invariant, which means they share the same network parameters across different parameterizations of the underlying functional data. In recent times, a lot of research is being carried out in the domain of operator regression \cite{goswami2022physics,lin2021operator}. To study the benefits of over-parameterization of DNNs, we have implemented two operator networks that have shown promising results so far, the DeepONet \cite{lu2021learning} and the Fourier neural operator (FNO) \cite{li2020fourier}. Although the original DeepONet architecture proposed in \cite{lu2021learning} has shown remarkable success, several extensions have been proposed in \cite{lu2021comprehensive} to modify its implementation and produce  efficient and robust architectures. In this section, we first provide an overview of the DeepONet framework put forth in \cite{lu2021learning}. This is followed by a proposed extension of the DeepONet, \textit{DeepONet with self-adaptivity}, to enhance the prediction capability of the network for non-smooth problems. We then briefly review the POD-DeepONet and FNO, which we use in this work. 

\subsubsection{DeepONet}

The DeepONet architecture consists of two DNNs: one encodes the input function at fixed sensor points (branch net), while another encodes the information related to the spatio-temporal coordinates of the output function (trunk net). 
The goal of the DeepONet is to learn the solution operator, $\mathcal G(\mathbf{x})$ that approximates $\mathcal{M}(\mathbf{x})$, and can be evaluated at continuous spatio-temporal coordinates, $\zeta$ (input to the trunk net). The output of the DeepONet for a specified input vector, $\mathbf{x}_i$, is a scalar-valued function of $\zeta$ expressed as $\mathcal G_{\bm\theta}(\mathbf{x}_i)(\zeta)$, where $\bm{\theta} = \left(\mathbf W, \mathbf b \right)$ includes the trainable parameters (weights, $\mathbf W$, and biases, $\mathbf b$) of the networks. 

\begin{figure}[ht!]
\begin{center}
\includegraphics[width=1\textwidth]{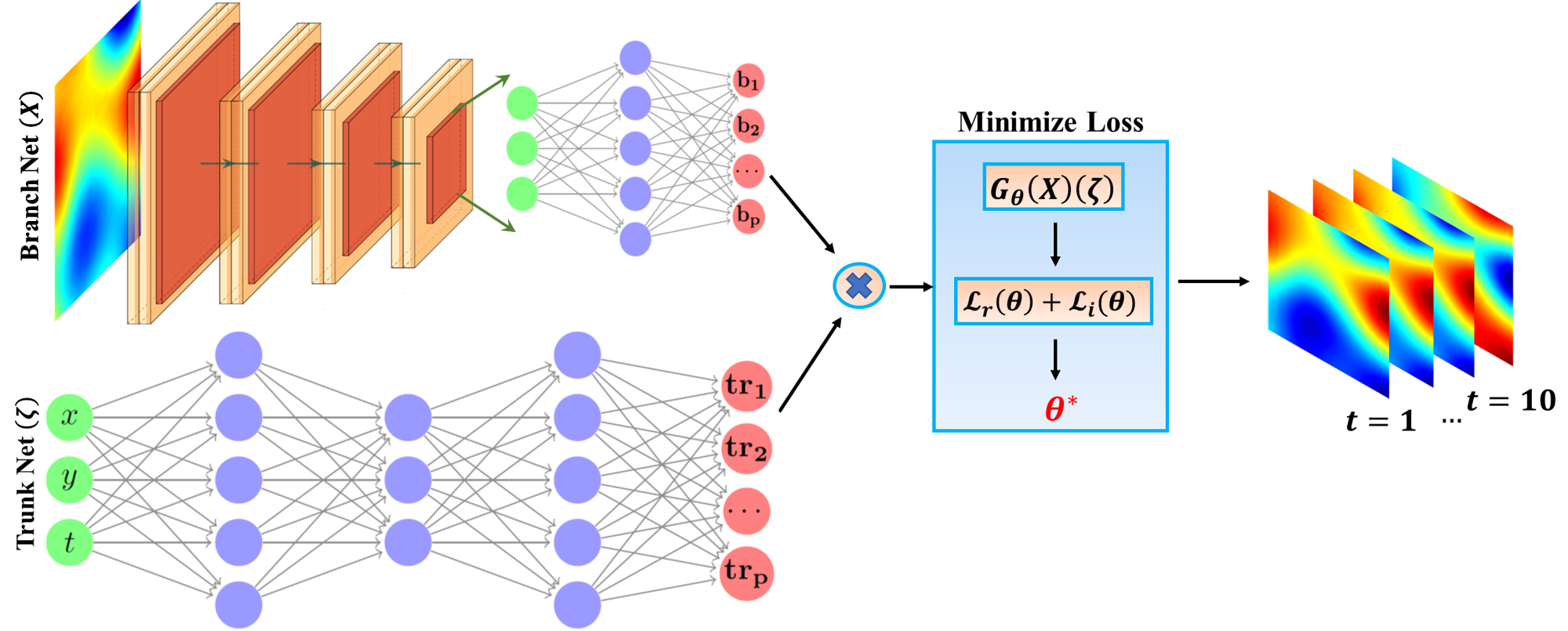}
\caption{Approximating PDE dynamics using the standard DeepONet. The DeepONet aims to learn the mapping between and initial random field (top left) and the time evolving solution (right). The schematic representation of the DeepONet considers a CNN architecture in the branch net, which takes as input the initial random field with training data $\mathbf X = \{\mathbf x_1, \mathbf x_2, \ldots, \mathbf x_N\},$ where $\mathbf x_i \in \mathbb{R}^{D_{in}}$, while the trunk net is shown with a FNN architecture that inputs the spatial and temporal coordinates, $[x,y,t]\in \zeta$. The dot product of the output feature embedding of the branch net, $[b_1, b_2, \ldots, b_p]^T \in \mathbb{R}^p$, and the trunk net, $[tr_1, tr_2, \ldots, tr_p]^T \in \mathbb{R}^p$, yields the solution operator, $\mathcal G_\theta$, where $\theta$ denotes the trainable parameters of the network. The loss function, $\mathcal L(\theta)$ obtained as a sum of the residual loss, $\mathcal L_r(\theta)$, and the initial condition loss, $\mathcal L_i(\theta)$, is minimized to obtain the optimized parameters of the network, $\theta^{*}$. The trained DeepONet is then used to predict the solution at any spatial and temporal location.}
\label{fig:DeepONet}
\end{center}
\end{figure}

In general, the input to the branch net is flexible, i.e., it can take the shape of the physical domain, the initial or boundary conditions, constant or variable coefficients, source terms, etc. But it must be discretized to represent the input function, $\mathbf X$ in a finite-dimensional space using a finite number of points, termed as sensors. We specifically evaluate $\mathbf X$ at fixed sensor locations $\{x_1, x_2, \dots, x_m\}$ to obtain the pointwise evaluations, $\mathbf{x}_i=\{\mathbf{x}_{i}(x_1), \mathbf{x}_{i}(x_2), \ldots, \mathbf {x}_{i}(x_m)\}$, which are used as input to the branch net. The trunk net takes as input the spatial and the temporal coordinates, e.g. $\zeta = \{x_i, y_i, t_i\}$, at which the solution operator is evaluated to compute the loss function. The solution operator for an input realization, $\mathbf x_1$, can be expressed as: 
\begin{equation}\label{eq:output_deeponets}
    \begin{split}
      \mathcal G_{\bm \theta}(\mathbf{x}_1)(\zeta) &= \sum_{i = 1}^p b_i \cdot tr_i = \sum_{i = 1}^{p}b_i(\mathbf{x}_{1}(x_1), \mathbf{x}_{1}(x_2), \ldots, \mathbf {x}_{1}(x_m))\cdot tr_i(\zeta),   
\end{split}
\end{equation}
where ${b_1, b_2, \ldots, b_p}$ are outputs of the branch net and ${tr_1, tr_2, \ldots, tr_p}$ are outputs of the trunk net. Conventionally, the trainable parameters of the DeepONet, represented by $\bm{\theta}$ in Eq.~\eqref{eq:output_deeponets}, are obtained by minimizing a loss function, which is expressed as:
\begin{equation}
    \mathcal L(\bm{\theta}) = \mathcal L_r(\bm{\theta}) + \mathcal L_i(\bm{\theta}),
\end{equation}
where $\mathcal L_r(\bm{\theta})$ and $\mathcal L_i(\bm{\theta})$ denote the residual loss and the initial condition loss, respectively. 

The DeepONet model provides a flexible paradigm that does not limit the branch and trunk networks to any particular architecture. For an equispaced discretization of the input function (used here), a Convolutional Neural Network (CNN) could be used for the branch net architecture. For a sparse representation of the input function, one could also use a Feedforward Neural Network (FNN). A standard practice is to use a FNN for the trunk network to take advantage of the low dimensions of the evaluation points, $\zeta$. A schematic representation of the standard DeepONet used to approximate the Brusselator dynamics described below is shown in Figure \ref{fig:DeepONet}, where the branch net is considered as a CNN, and the trunk net has a FNN architecture.

DeepONet aims to infer a continuous latent function that arises as the solution to a system of nonlinear PDEs. Neural networks employ back-propagation algorithm to tune the network parameters, while trying to reflect the best fit solution to the training data. 
Approximating steep gradients is often challenging for neural networks. Efforts in the literature aim to impose such constraints in a soft manner by appropriately penalizing the loss function of CNN approximations. However, manually manoeuvring for the optimal penalizing parameter is a tedious task and often leads to unstable and erroneous predictions. We propose a simple solution to this problem, \emph{DeepONet with self-adaptivity}, where the penalty parameters are trainable, so the neural network learns by itself which regions of the solution are difficult and is forced to focus on them.
\\
\\
\noindent
\textbf{A.} \textit{DeepONet with Self-adaptivity}

The basic idea behind self-adaptivity is to make the penalty parameters increase where the corresponding loss is higher, which is accomplished by training the network to simultaneously minimize the losses and maximize the value of the penalty parameters. Discontinuities or non-smooth features in the solution lead to a non-differentiable loss function at these locations. Hence, even though the overall error is reduced during the training, there is significant error near the discontinuity. The conventional approach to take care of such specific local minimization issues is to introduce constant penalty terms to force the network to satisfy the conditions. Accordingly, the loss function would be of the form:
\begin{equation}
    \mathcal L(\boldsymbol{\theta}) = \mathcal L_r(\boldsymbol{\theta}) + \mathcal L_i(\boldsymbol{\theta}) + \gamma_1\mathcal L_{d1}(\boldsymbol{\theta}) + \gamma_2\mathcal L_{d2}(\boldsymbol{\theta}),
\end{equation}
where $\gamma_1 >>1$ and $\gamma_2>>1$ are manually chosen constant parameters, and $\mathcal L_{d1}(\boldsymbol{\theta})$ and $\mathcal L_{d2}(\boldsymbol{\theta})$ are location-specific losses corresponding to the location of the discontinuity. The optimal values for $\gamma_1$ and $\gamma_2$ differ widely for different PDEs. Considering that in a DeepONet we are trying to find the solution operator for multiple PDEs by training the network just once, obtaining an optimal penalty parameter is a challenge.

In this section, we introduce a simple solution to tune these parameters: make these parameters trainable. These hyper-parameters are updated by back-propagation together with the network weights. The new loss function is therefore expressed as:
\begin{equation}
    \mathcal L(\boldsymbol{\theta},\lambda_1, \lambda_2) = \mathcal L_r(\boldsymbol{\theta}) + \mathcal L_i(\boldsymbol{\theta}) + \lambda_1\mathcal L_{d1}(\boldsymbol{\theta},\gamma_1) + \lambda_2\mathcal L_{d2}(\boldsymbol{\theta}, \gamma_2),
\end{equation}
where $\lambda_1, \lambda_2$ are self-adaptive hyperparameters. Typically, in a neural network, we minimize the loss function with respect to the network parameters, $\boldsymbol{\theta}$. However, in this approach we additionally maximize the loss function with respect to the trainable hyperparameters using a gradient descent/ascent procedure. The proposed approach shares the same concept of introducing trainable hyperparameters as in \cite{mcclenny2020self}.
\\
\\
\noindent
\textbf{B.} \textit{POD-DeepONet}

In this work, we use an existing extension of the standard DeepONet proposed in \cite{lu2021comprehensive} referred to as the POD-DeepONet. The standard DeepONet employs the trunk net to learn the basis of the output function from the data. However, in POD-DeepONet, the basis functions are computed by performing proper orthogonal decomposition (POD) on the training data (after the mean has been excluded), and using this basis in place of the trunk net. A deep neural network is employed in the branch net to learn the POD basis coefficients such that the output can be written as:
\begin{equation}
    \mathcal{G}(\mathbf x)(\zeta) = \sum_{i=1}^p b_i(\mathbf x) \phi_i(\zeta) + \phi_0(\zeta)
\end{equation}
where $\phi_0(\zeta)$ is the mean function of all $\mathbf x_i(\xi), i =1,\dots,N$ computed from the training dataset, and $\{\phi_1, \phi_2, \dots, \phi_p\}$ are the $p$ precomputed POD modes of $\mathbf x_i(\zeta)$. 


\subsubsection{Fourier Neural Operator}

The Fourier Neural Operator (FNO) is based on parameterizing the integral kernel in the Fourier space. The concept was initially proposed in \cite{li2020fourier}. FNO in its continuous form can be conceptualized as a DeepONet with a particular network architecture of the branch net and the trunk net expressed by discrete trigonometric basis functions. Neural networks are typically trained to approximate functions that are described in the Euclidean space, the conventional graph with spatial coordinate axes. However, in FNO, the network parameters, inputs to the network and the output from the network are specified in the Fourier space, a distinct type of graph used to represent frequencies.

FNO employs evaluations restricted on an equispaced mesh to discretize both the input and output functions, where the mesh and the domain must be the same. Simplistically, the input random field, $\mathbf X$, the evaluation locations, $\zeta$, and the output function, $\mathbf Y$, are defined on the same domain with the same equispaced discretization.  The input to the network is mapped on a linear layer, converting the input to a higher dimension, followed by passing it through four layers of integral operators and later projecting it back to the original dimension. Within each integral operator operation, there is a additional linear and nonlinear transform where the Fourier layers act as low-pass spectral filters. In practice, the retention of more Fourier modes (maximum is half the resolution+1) has helped in achieving better accuracy. Details on the implementation of FNO can be found in \cite{li2020fourier}.

\subsection{Manifold learning based surrogates}
\label{sec:manifold-PCE}
Manifold learning based surrogates are used to project high-dimensional input-output data from a mathematical model onto a low-dimensional manifold where interpolation is performed for rapid solution prediction. The manifold-based polynomial chaos expansion or manifold PCE \textit{(m-PCE}), specifically aims to construct a PCE surrogate model to map the reduced input data to the corresponding QoIs (also potentially in a reduced space), thus mitigating the curse of dimensionality. A detailed survey on the use of various dimension reduction (DR) techniques for the construction of \textit{m-PCE} surrogates was conducted by Kontolati et al.\ \cite{kontolati2022survey}. 

In the following sections, we introduce the standard approach for constructing PCE surrogates and then present the k-PCA PCE, a specific implementation of \textit{m-PCE} that leverages kernel-Principal Component Analysis (k-PCA) for dimension reduction.  Despite the robust performance of \textit{m-PCE} as demonstrated in \cite{kontolati2022survey}, in certain cases where QoIs are very high dimensional, the accuracy of \textit{m-PCE} surrogates may diminish. In the following, we propose an extension of the \textit{m-PCE} framework in which a manifold projection based DR is performed on both high-dimensional inputs and high-dimensional solutions.

\subsubsection{Standard PCE}
We assume the model $\mathcal{M}\left(\mathbf{x}\right)$, where $\mathbf{x}$ is a $k$-variate random variable defined on the probability space $\left(\Omega, \Sigma, P\right)$ and characterized by the joint probability density function (PDF) $\varrho_{\mathbf{x}}: X \rightarrow \mathbb{R}_{\geq 0}$, where $X \subseteq \mathbb{R}^k$ is the image space, $\Omega$ the sample space, $\Sigma$ the set of events, and $P$ the probability measure.
By further assuming that $\mathcal{M}$ satisfies the Doob-Dynkin lemma \cite{bobrowski2005functional}, its output $\mathcal{M}\left(\mathbf{x}\right)$ is a random variable dependent on $\mathbf{x}$.
In the following, we consider for simplicity a scalar output $Y = \mathcal{M}\left(\mathbf{x}\right) \in \mathbb{R}$ although the extension to multivariate outputs is straightforward by applying the PCE approximation element-wise. 
The PCE is a spectral approximation of the form
\begin{equation}
\label{eq:spectral_approx}
\mathcal{M}(\mathbf{x}) \approx \widetilde{\mathcal{M}}(\mathbf{x}) = \sum_{s=1}^S c_s \Xi_s(\mathbf{x}),
\end{equation}
where $c_s$ are scalar coefficients and $\Xi_s$ are multivariate polynomials that are orthonormal with respect to the joint PDF $\varrho_{\mathbf{x}}$, such that
\begin{equation}
\label{eq:orthNd}
\mathbb{E}\left[\Xi_s \Xi_t\right] = \int_{Z}  \Xi_s\left(\mathbf{x}\right) \Xi_t\left(\mathbf{x}\right) \varrho_{\mathbf{x}}\left(\mathbf{x}\right) \mathrm{d}\mathbf{x} =  \delta_{st},
\end{equation}
where $\delta_{st}$ denotes the Kronecker delta.
Depending on the PDF $\varrho_{\mathbf{x}}$, the orthonormal polynomials can be chosen according to the Wiener-Askey scheme \cite{xiu2002wiener} or constructed numerically\cite{wan2006multi, soize2004physical}.
Since $\mathbf{x}$ is assumed to consist of independent random variables $x_1, \dots, x_k$, the joint PDF is given as 
\begin{equation}
\label{eq:joint_pdf}
\varrho_{\mathbf{x}} \left(\mathbf{x}\right) = \prod_{i=1}^k \varrho_{x_i}\left(x_i\right),
\end{equation}
where $\varrho_{x_i}$ is the marginal PDF of random variable $x_i$.
Likewise, the multivariate orthogonal polynomials are constructed as 
\begin{equation}
\label{eq:polyNd}
\Xi_s(\mathbf{x}) \equiv \Xi_\mathbf{s}(\mathbf{x}) = \prod_{i=1}^k \xi_i^{s_i} (x_i),
\end{equation}
where $\xi_i^{s_i}$ are univariate polynomials of degree $s_i \in \mathbb{Z}_{\geq 0}$ and are orthonormal with respect to the univariate PDF $\varrho_{x_i}$, such that
\begin{equation}
\label{eq:orth1d}
\mathbb{E}\left[\xi_i^{s_i}\xi_i^{t_i}\right] = \int_{Z_i} \xi_i^{s_i}(x_i) \xi_i^{t_i}(x_i) \varrho_{x_i}(x_i) \, \mathrm{d}X_i = \delta_{s_i t_i}. 
\end{equation}
The multi-index $\mathbf{s} = \left(s_1, \dots, s_k\right)$ is equivalent to the multivariate polynomial degree and uniquely associated to the single index $s$ employed in Eq.~\eqref{eq:spectral_approx}, which can now be written in the equivalent form
\begin{equation}
\label{eq:spectral_approx_multi_index}
\mathcal{M}(\mathbf{x}) \approx \widetilde{\mathcal{M}}(\mathbf{x}) = \sum_{\mathbf{s} \in \Lambda} c_{\mathbf{s}} \Xi_{\mathbf{s}}(\mathbf{x}),
\end{equation}
where $\Lambda$ is a multi-index set with cardinality $\#\Lambda = S$.
The choice of the multi-index set $\Lambda$ plays a central role in the construction of the PCE, as it defines which polynomials and corresponding coefficients form the PCE.
The most common choice, as well as the one employed in this work, is that of a total-degree multi-index set, such that $\Lambda$ includes all multi-indices that satisfy $\left\| \mathbf{s} \right\|_1 \leq s_{\max}$, $s_{\max} \in \mathbb{Z}_{\geq 0}$. In that case, the size of the PCE basis is $S = \frac{\left(s_{\max} + k\right)!}{s_{\max}! k!}$, such that it scales polynomially with the input dimension $k$ and the maximum degree $s_{\max}$. 

Several approaches have been proposed for computing the coefficients $c_{\mathbf{s}}$, including pseudo-spectral projection \cite{constantine2012sparse, conrad2013adaptive, winokur2016sparse}, interpolation \cite{buzzard2013efficient, loukrezis2019adaptive}, and, most commonly, regression \cite{blatman2011adaptive, loukrezis2020robust, hampton2018basis, diaz2018sparse, hadigol2018least, he2020adaptive, tsilifis2019compressive}. We employ a regression approach in which the PCE coefficients are determined by solving the penalized least squares problem \cite{rifkin2007notes}  
\begin{align}
\label{eq:regression}
\underset {\mathbf{c} \in \mathbb{R}^{\#\Lambda}}{\arg\min} \left\{\frac{1}{N}\sum_{i=1}^{N} \left( \mathcal{M}(\mathbf{x}_i) - \sum_{\mathbf{s} \in \Lambda} c_{\mathbf{s}} \Xi_{\mathbf{s}}\left(\mathbf{x}_i\right) \right)^2 + \lambda J\left(\mathbf{c}\right)\right\},
\end{align}
where $\lambda \in \mathbb{R}$ is a penalty factor, $J\left(\mathbf{c}\right)$ a penalty function acting on the vector of PCE coefficients $\mathbf{c} \in \mathbb{R}^{\#\Lambda}$, and $\mathbf{X} = \left\{\mathbf{x}_i\right\}_{i=1}^{N}$ is an experimental design (ED) containing realizations of $\mathbf{x}$ with corresponding model outputs $\mathbf{Y} = \left\{\mathbf{y}_i\right\}_{i=1}^{N}$. Common choices for the penalty function $J(\mathbf{c})$ are the $\ell_1$ and $\ell_2$ norms, in which cases problem~\eqref{eq:regression} is referred to as LASSO (least absolute shrinkage and selection operator) and ridge regression, respectively. Removing the penalty term results in an ordinary least squares regression problem.

\subsubsection{k-PCA PCE}

Manifold PCE employs manifold learning techniques for DR to identify lower-dimensional embeddings for the construction of efficient and accurate PCE surrogates. DR allows us to neglect redundant features and noise, therefore avoiding overfitting and tedious training. The method is based on a two-step approach: 1) the lower-dimensional manifold is discovered from the input and/or output data and 2) a PCE surrogate is constructed to approximate the map between reduced inputs and outputs based on a limited set of training data. In general, any DR method can be employed, including linear methods like principal component analysis (PCA) and nonlinear methods such as independent component analysis (ICA), diffusion maps (DMAPs), non-negative matrix factorization (NMF), autoencoders etc. In this work, we employ k-PCA for DR.

\begin{figure}[ht!]
\begin{center}
\includegraphics[width=1\textwidth]{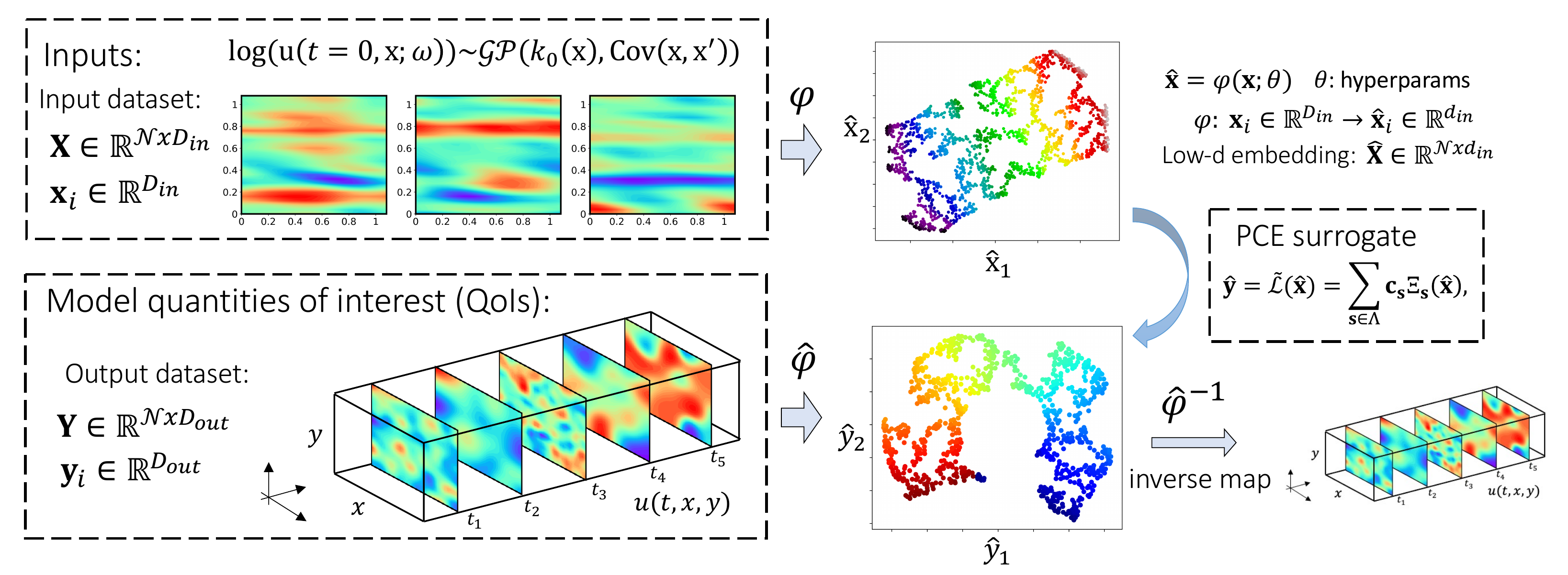}
\caption{Schematic of the proposed \textit{m-PCE} approach. Left to right: Lower-dimensional embeddings are identified for both inputs $\mathbf{X} = \{\mathbf{x}_{1},..,\mathbf{x}_{N}\}$ (top) and outputs $\mathbf{Y} = \{\mathbf{y}_{1},..,\mathbf{y}_{N}\}$ (bottom) via two mappings $\varphi$ and $\hat{\varphi}$ respectively. A PCE surrogate $\tilde{\mathcal{L}}$ is constructed to map reduced inputs $\hat{\mathbf{x}}$ to reduced outputs $\hat{\mathbf{y}}$. Finally, the inverse transformation $\hat{\varphi}^{-1}$ is applied to project low-dimensional predictions back to the original space.}
\label{fig:manifold-PCE-schematic}
\end{center}
\end{figure}

In the original development of  \textit{m-PCE}\cite{kontolati2022survey}, DR is performed only on the high-dimensional input. In this work, we extend the framework to include DR on the output leveraging the method developed in \cite{kontolati2021manifold}. We note that, in cases where the output data undergo a reduction, the DR method must possess an inverse transformation to map predicted outputs in the low-dimensional space to the physically interpretable ambient space. This proposed framework is illustrated in Figure \ref{fig:manifold-PCE-schematic} and described in detail in the following where first we describe the original \textit{m-PCE} with DR only on the input and then describe the proposed methods with DR on both input and output. For completeness, we then present the k-PCA method employed for DR in this work.

\noindent
\textbf{Original m-PCE:} \textit{DR on input only}\\
In the original \textit{m-PCE}, the method begins with an input training data set
$\mathbf{X} = \{\mathbf{x}_{1},..,\mathbf{x}_{N}\}$ where $\mathbf{x}_i \in \mathbb{R}^{D_{\text{in}}}$. We assume that the intrinsic dimension of the data set $d_{\text{in}} \ll D_{\text{in}}$. Therefore, the input data lie on or near a manifold of dimensionality $d_{\text{in}}$ that is embedded in the $D_{\text{in}}$-dimensional space. We identify a low-dimensional representation of the original dataset $\hat{\mathbf{X}} = \{\hat{\mathbf{x}}_{1},..,\hat{\mathbf{x}_{N}}\}$, $\hat{\mathbf{x}}_i \in \mathbb{R}^{d_{in}}$ by constructing a mapping $\hat{\mathbf{x}}=\varphi(\mathbf{x};\theta)$, where $\varphi: \mathbf{x}_i \in \mathbb{R}^{D_{in}} \rightarrow \hat{\mathbf{x}}_i \in \mathbb{R}^{d_{in}}$ and $\theta \in \mathbb{R}^n$ represents the $n$-dimensional vector of parameters associated with the reduction method. The DR method is learned by identifying the parameters $\theta$ that minimize an error measure, $L(\mathbf{x};\theta)$, as 
\begin{equation}
    \hat{\theta} = \underset {\theta}{\arg\min} L(\mathbf{x};\theta).
\end{equation}
For example, when the inverse transformation $\varphi^{-1}$ is available and the objective is to reduce the information loss after the data reduction, the error measure can be represented by the mean-square reconstruction error \cite{lataniotis2020extending}.

Using the reduced representation of the input, $\mathbf{x}$, we then construct a PCE surrogate model $\mathbf{y}=\widetilde{\mathcal{M}}(\hat{\mathbf{x}})$ as described in Eq.\ \eqref{eq:spectral_approx_multi_index}. Given that $\mathbf{y}$ is not generally scalar-valued, PCEs are constructed component-wise as $y_i=\widetilde{\mathcal{M}}_i(\hat{\mathbf{x}})$. For prediction purposes at a new point $\mathbf{x}^*$, the point is first projected onto the low-dimensional manifold by $\hat{\mathbf{x}}^*=\varphi(\mathbf{x}^*)$ and the PCE is used on this projected point to predict the solution. For additional details, see \cite{kontolati2022survey}.
\\
\\
\textbf{Proposed m-PCE:} \textit{DR on both input and output}\\
Commonly, in physics-based models, QoIs are evaluated at many points in space and time resulting in very high-dimensional ($D_{out}$) output which standard surrogate modeling techniques cannot handle without loss in predictive accuracy. Assuming that the intrinsic dimension of this output data $d_{out}\ll D_{out}$, we propose to use DR on both the input and output data as a preprocessing step prior to the construction of the PCE surrogate. With respect to the input data, the method proceeds exactly as in the previous section. However, we introduce a second mapping $\hat{\mathbf{y}}=\hat{\varphi}(\mathbf{y};\hat{\theta})$, where $\hat{\varphi}: \mathbf{y}_i \in \mathbb{R}^{D_{\text{out}}} \rightarrow \hat{\mathbf{y}}_i \in \mathbb{R}^{d_{\text{out}}}$ and $\hat{\theta} \in \mathbb{R}^{\hat{n}}$ that serves to reduce the dimension of the output data $\mathbf{Y} = \{\mathbf{y}_{1},..,\mathbf{y}_{N}\}$. Here, the DR technique employed must possess an inverse map $\hat{\phi}^{-1}: \hat{\mathbf{y}}_i \in \mathbb{R}^{d_{\text{out}}} \rightarrow \mathbf{y}_i \in \mathbb{R}^{D_{\text{out}}} $, which will act as a decoder for projecting low-dimensional outputs back to the ambient space. 

Once the two embeddings have been identified, a PCE surrogate $\widetilde{\mathcal{M}}$ is constructed to map the reduced inputs to the outputs as $\widetilde{\mathcal{M}}: \hat{\mathbf{x}}_i \in \mathbb{R}^{d_{\text{in}}} \rightarrow  \hat{\mathbf{y}}_i \in \mathbb{R}^{d_{\text{out}}}$. For prediction purposes, the reduced input at a new point $\hat{\mathbf{x}}^*$ is used to predict the reduced solution $\hat{\mathbf{y}}^*=\widetilde{\mathcal{M}}(\hat{\mathbf{x}}^*)$, which is then project back to the ambient space by $\mathbf{y}^*=\hat{\varphi}^{-1}(\hat{\mathbf{y}}^*)$. To evaluate the accuracy of the proposed approach, data generated by the PCE surrogate are transformed back to the original space $\mathbb{R}^{D_{\text{out}}}$, and compared with the ground truth. This proposed \textit{m-PCE} framework is presented in Figure \ref{fig:manifold-PCE-schematic}. 
\\
\\
\noindent
\textbf{DR with k-PCA:}

In this work, we use k-PCA to identify the lower-dimensional embeddings of both the inputs and the outputs. In this section, we present the method generically and we use $D$ to denote the original dimension of the data, which may refer to either $D_{\text{in}}$ or $D_{\text{out}}$. 

k-PCA \cite{scholkopf1997kernel, hoffmann2007kernel} is the nonlinear variant of standard PCA and extends the method to nonlinear data distributions. In k-PCA, data points are mapped onto a higher-dimensional feature space $\mathcal{F}$, as
\begin{equation}
\label{eq:map}
    \mathbf{x}_i \rightarrow \mathbf{\Phi}(\mathbf{x}_i),
\end{equation}
where $\mathbf{\Phi}: \mathbb{R}^D \rightarrow \mathbb{R}^N$ with $N>D$. Standard PCA is then performed in this higher-dimensional space. The above transformation is obtained implicitly through the use of a kernel function $k(\mathbf{x}_i, \mathbf{x}_j)$ which replaces the scalar product using the relation $k(\mathbf{x}_i, \mathbf{x}_j) = \mathbf{\Phi}(\mathbf{x}_i)^{\text{T}} \mathbf{\Phi}(\mathbf{x}_j)$, thus $\mathbf{\Phi}$ is never calculated explicitly. This process is known as \emph{`kernel trick'} and allows the transformation of datasets to very high-dimensional spaces, therefore allowing the encoding of highly nonlinear manifolds without explicit knowledge of suitable feature functions \cite{bishop2006pattern}. Since data in the featured space are not guaranteed to have zero-mean, the kernel matrix is centralized as
\begin{equation}
\label{eq:center}
    \mathbf{K}' = \mathbf{K} - \mathbf{1}_N \mathbf{K} - \mathbf{K} \mathbf{1}_N + \mathbf{1}_N \mathbf{K}, \mathbf{1}_N,
\end{equation}
where $\mathbf{1}_N$ represents an $N \times N$ matrix with values $1/N$ and $\mathbf{K}$ is the kernel matrix.

The k-PCA formulation does not compute the principal components (PCs) directly, but rather it projects the data onto the PCs. The projection of a mapped data point $\mathbf{\Phi}(\mathbf{x}_i)$ onto the $k$-th PC $\mathbf{W}^k$ in $\mathcal{F}$ is computed as
\begin{equation}
\label{eq:kernel-pca}
    (\mathbf{W}^k)^{\text{T}} \cdot \mathbf{\Phi}(\mathbf{x}) = \bigg(\sum_{i=1}^{n}\alpha_i^k \mathbf{\Phi}(\mathbf{x}_i) \bigg)^{\text{T}} \mathbf{\Phi}(\mathbf{x}),
\end{equation}
where again $\mathbf{\Phi}(\mathbf{x}_i)^{\text{T}}\mathbf{\Phi}(\mathbf{x})$ is the dot product obtained from the entries of $\mathbf{K}'$. The coefficients $\alpha_i^k$, are determined by solving the eigenvalue problem
\begin{equation}
\label{eq:eigen}
    N \lambda \boldsymbol{\alpha} = \mathbf{K}' \boldsymbol{\alpha},
\end{equation}
where $\lambda$ and $\boldsymbol{\alpha}$ are the eigenvalues and eigenvectors of $\mathbf{K}'$ respectively and $N$ is the number of data points. Application of the method is dependent on the selection of a positive semi-definite kernel function, of which there are many options including linear, polynomial, and Gaussian kernels.

\section{Brusselator reaction-diffusion system}
\label{sec:bruss}

As a prototypical physico-chemical system, we consider the Brusselator diffusion-reaction system introduced by Ilya Prigogine in the 1970s \cite{prigogine1978time}, which describes an autocatalytic chemical reaction in which a reactant substance interacts with another substance to increase its production rate \cite{ahmed2019numerical}. The system is selected because it is mathematically described by a set of PDEs that exhibit nonlinear features that are particularly difficult for approximate models to replicate. The Brusselator model is characterized by the following reactions
\begin{subequations}\label{eq:reaction_model}
\begin{align}
    A & \xrightarrow{k_1} X, \quad \\ 
    B + X  & \xrightarrow{k_2} Y + D, \\
    2X + Y & \xrightarrow{k_3} 3X,\\
    X & \xrightarrow{k_4} E,
\end{align}
\label{eq:A3-reaction}
\end{subequations}
where $k_i (i= 1, 2, 3, 4)$ are positive parameters representing the reaction rate constant. In Eq.~\eqref{eq:reaction_model}, a reactant, $A$ is converted to a final product $E$, in four steps with the help of four additional species, $X,B,Y,$ and $D$. We consider that $A$ and $B$ are in vast excess and thus can be modeled at constant concentration. The 2D rate equations becomes:
\begin{equation}
\begin{split}
    \frac{\partial u}{\partial t} &= D_0 \bigg(\frac{\partial^2 u}{\partial x^2} + \frac{\partial^2 u}{\partial y^2} \bigg) + a - (1+b)u + vu^2, \\
    \frac{\partial v}{\partial t} &= D_1 \bigg(\frac{\partial^2 v}{\partial x^2} + \frac{\partial^2 v}{\partial y^2} \bigg) + bu - vu^2, \hspace{10pt} \mathsf{x} \in [0,1]^2, \ t \in [0,1],
\end{split} 
\label{eq:A3-model}
\end{equation}
with the given initial conditions
\begin{align*}
u(x,y,t=0) &= h_1(x,y) \ge 0, \\
v(x,y,t=0) &= h_2(x,y) \ge 0,
\end{align*}
where $\mathsf{x}=(x,y)$ are the spatial coordinates, $D_0, D_1$ represent the diffusion coefficients, $a=\{A\}, b=\{B\}$ are constant concentrations, and $u=\{X\}, v=\{Y\}$ represent the concentrations of reactant species $X,Y$..

\subsection{Data generation}
\label{data-generation}
\begin{figure}[ht!]
\begin{center}
\includegraphics[width=1\textwidth]{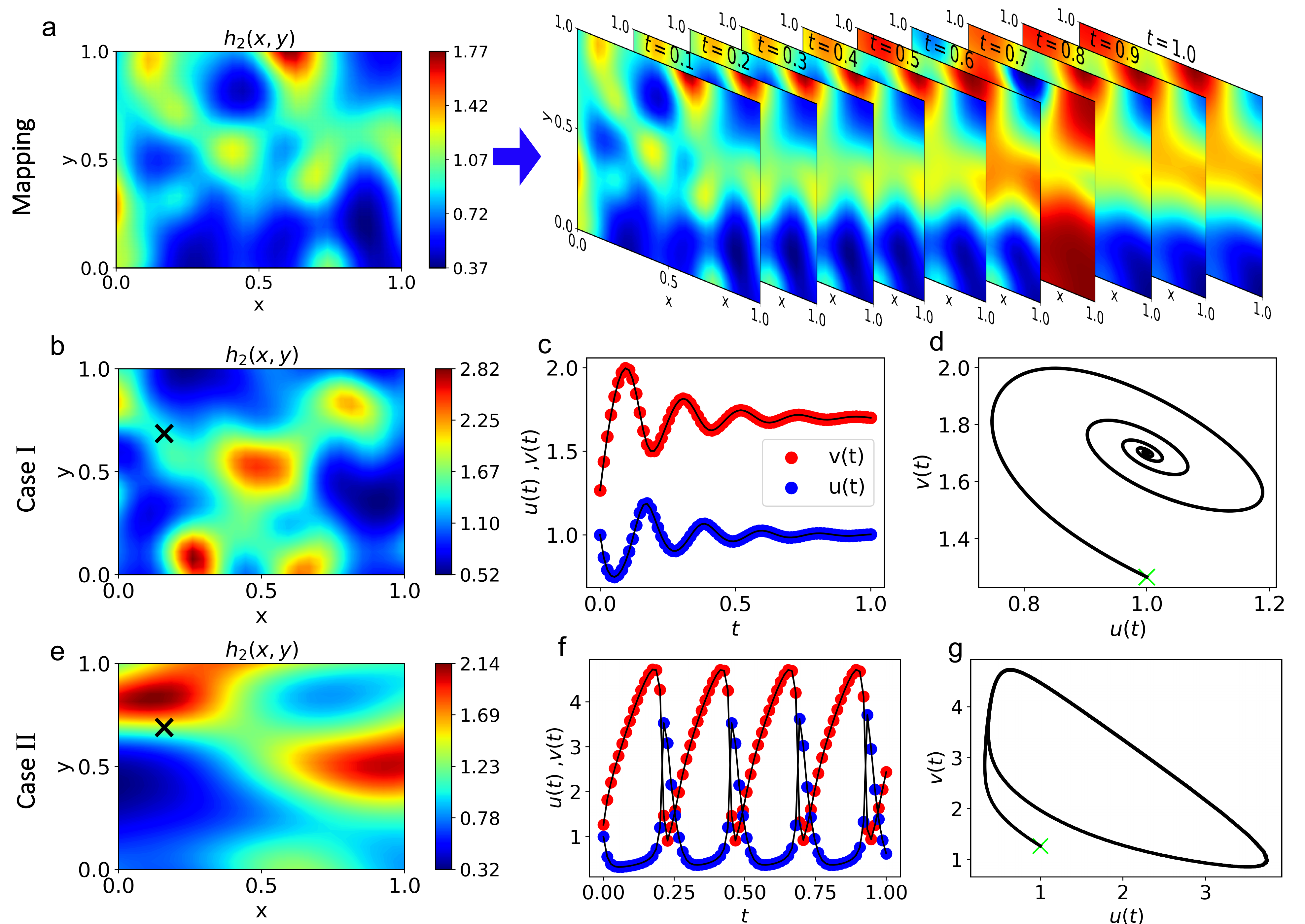}
\caption{Brusselator dynamics: (a) shows the mapping we aim to construct between initial random field $h_2(x,y)$ and $v(x,y,t)$ evaluated at 10 time steps for one full oscillation (b) represents the initial random field for Case I, where $b=1.7$, $l_x=0.11$, $l_y=0.15$, $\sigma^2=0.15$. (c) shows the trajectory of concentrations $u(t), v(t)$ at a fixed spatial point $(x,y)=(0.148,0.704)$ approaching a stable equilibrium, and (d) shows the corresponding model evolution in phase space. In (e), the initial random field is shown for Case II, where $b=3.0$, $l_x=0.35$, $l_y=0.20$, $\sigma^2=0.15$. (f) shows the trajectory of concentrations $u(t), v(t)$ at a fixed spatial point $(x,y)=(0.148,0.704)$ exhibits periodic oscillations, and (g) shows the corresponding limit cycle in the phase space. We note that we aim to map input random fields to snapshots corresponding to one full oscillation (first row), however in (c) and (f) we show the evolution of the system past one oscillation.}
\label{fig:brusselator-dynamics}
\end{center}
\end{figure}

We aim to learn the mapping from initial concentration $h_2(x,y)$ to the evolved concentration $v(x,y,t)$, where $t>0$ (as shown in Figure \ref{fig:brusselator-dynamics}(a)). The initial concentration $h_2(x,y)$ is modeled as a Gaussian random field
\begin{equation}
\label{eq:A1:normal}
    h_2(\mathsf{x}) \sim \mathcal{GP}(h_2(\mathsf{x})|\mu(\mathsf{x}), \text{Cov}(\mathsf{x},\mathsf{x}')),
\end{equation}
where $\mu(\mathsf{x})$ and $\text{Cov}(\mathsf{x},\mathsf{x}')$ are the mean and covariance functions respectively. For simplicity, we set $\mu(\mathsf{x})=0$, while the covariance matrix is given by the squared exponential kernel as
\begin{equation}
\label{eq:A2:cov}
    \text{Cov}(\mathsf{x},\mathsf{x}') = \sigma^2 \text{exp} \Bigg( - \frac{\|x - x'\|^2_2}{2\ell_x^2} - \frac{\|y - y'\|^2_2}{2\ell_y^2}\Bigg),
\end{equation}
where $\ell_x$, $\ell_y$ are the correlation length scales along the $x$ and $y$ spatial directions, respectively. To generate realizations of the input stochastic field, we employ the truncated Karhunen-Lo\'eve expansion. 

We consider two cases corresponding to different initial concentrations of reactant $B$ as illustrated in Figure \ref{fig:brusselator-dynamics}. In Case I, the concentrations $u(t)$ and $v(t)$ both approach a stable equilibrium concentration. In Case II, the system reaches a limit cycle which causes periodic oscillations in the concentrations $u(t)$ and $v(t)$. In both cases, we develop three sets of data: a training/test data set with specified random field parameters and two OOD data sets with different random field parameters. These parameter sets are presented in Table \ref{table:datasets}. Datasets OOD$_1$ and OOD$_2$ correspond to lower and higher length scales respectively, compared to the reference training data, to test the extrapolation accuracy of the trained models.

\begin{table}[ht!]
\footnotesize
\caption{Model and input/output parameters for Case I and Case II of the Brusselator model. OOD$_1$ and OOD$_2$ correspond to data sets used to test the surrogate models in extrapolation.}
\centering
\begin{tabular}{c c c c c c c}
\toprule
 & \multirow{2}{*}{Data} & \multirow{2}{*}{$b=\{B\}$} & \multirow{2}{*}{$n_t$} & \multicolumn{3}{c}{KLE parameters} \\ \cmidrule(l){5-7} 
& & & & $l_x$ & $l_y$ & $\sigma^2$ \\
 \toprule
\multirow{3}{*}{Case I}  & train/test & \multirow{3}{*}{$1.7$} & \multirow{3}{*}{$20$} & $0.11$ & $0.15$ & $0.15$ \\
  & OOD$_1$  &  &   & $0.09$ & $0.20$ & $0.18$ \\ 
  & OOD$_2$ &  &  &  $0.35$ & $0.20$ & $0.15$ \\  \hdashline
\multirow{3}{*}{Case II}  & train/test & \multirow{3}{*}{$3.0$} & \multirow{3}{*}{$10$} & $0.35$ & $0.20$ & $0.15$ \\
  & OOD$_1$  &  &   & $0.25$ & $0.15$ & $0.15$ \\ 
  & OOD$_2$ &  &  &  $0.45$ & $0.40$ & $0.15$ \\ 
\bottomrule
\end{tabular}
\label{table:datasets}
\end{table}

In all cases, the initial field $h_1(x,y)$ is kept constant at $h_1(x,y) = a = 1$. The diffusion coefficients are set equal to $D_0=1$ and $D_1=0.5$. The simulation takes place in a square domain $\Omega= [0,1] \times [0,1]$, discretized with $28\times28=784$ grid points and solved with finite differences (FD) in the time interval $t=[0,1]$ for $\delta t=10^{-2}$. To form the training data set, we collect solution snapshots $v(x,y,t)$ at $n_t=20$ (Case I) and $n_t=10$ (Case II) equally spaced points in time. We denote the input random field with $\mathbf{X} = \{\mathbf{x}_{1},..,\mathbf{x}_{N}\}$, where $\mathbf{x}_i \in \mathbb{R}^{D_{\text{in}}}$, $D_{\text{in}}=784$ and corresponding model outputs $\mathbf{Y} = \{\mathbf{y}_{1},..,\mathbf{y}_{N}\}$, where $\mathbf{y}_i \in \mathbb{R}^{D_{\text{out}}}$, $D^{I}_{\text{out}}=20 \times D_{\text{in}}= 15,680$, $D^{II}_{\text{out}}=10 \times D_{\text{in}}= 7840$ and $N$ is the number of training data.


\subsection{Surrogate Approximations and their Parameterizations}

For the data sets described in the previous sections, we develop surrogate models using the DeepONet, FNO, and m-PCE methods. In this section, we briefly discuss some important aspects of these surrogates. 

For the DeepONet, we consider different architectures for the two cases. In Case I, we observe a smooth and decaying oscillation, which is encoded into the network simply by modifying the time input of the trunk network, as a trigonometric function, to obtain appropriate basis functions. However, for Case II, we observe that even though the function exhibits periodic oscillations, there are sudden spikes towards the end of each cycle. In this case, we use the proposed modified DeepONet with self-adaptivity (SA-DeepONet). The DeepONet implemented for this case uses a FNN architecture in both the branch and the trunk net. The spikes in the solution field observed toward the end of each cycle lead to a non-differentiable loss function, as seen in Figure \ref{fig:brusselator-dynamics}, which necessitates self-adaptivity. In FNO, we have used a equispaced grid discretization for the spatial and the temporal domain as input to the integral-operator based framework. 

For m-PCE surrogate modeling, we investigate both under-parameterized (U m-PCE) and over-parameterized (O m-PCE) versions of the model. In the U m-PCE, the total number of trainable PCE parameters is less than the number of available training data points (i.e. $n_p < N$) where
\begin{equation}
    n_p = \dfrac{(s_{max}+d_{in})!}{s_{max}!d_{in}!}d_{out}.
\end{equation}
Meanwhile, for the O m-PCE we have $n_p>N$. Note that $n_p$ depends on the PCE order as well as the dimensions of the input/output latent spaces. Thus, increasing any of these quantities will increase model complexity.  We further note that we use standard PCE learning methods with full index sets and low-order ($s_{\max}\le 4$) polynomials. We do not consider sparse PCE implementations that use, for example Least Angle Regression \cite{blatman2011adaptive}, to reduce the basis set and promote sparsity in the expansion. This is consistent with the DNN architecture we use where, likewise, no attempt is made to explore alternative or more compact architectures.

\section{Numerical results}
\label{sec:results}

Here, we explore the performance of manifold PCE, DeepONet, and FNO on the Brusselator system. To evaluate the performance of the model, we compute the $L_2$ relative error of predictions, and we report the mean and standard deviation of this metric based on five independent training trials. Below, we present the results for the two cases of the model discussed in Section \ref{data-generation}.

\subsection{Case I}
\label{CaseI}

Initially, we train all models with $N=800\times n_t=16,000$ data and test their accuracy on a test data sets of $N_*=200\times n_t=4,000$ data. In Table \ref{table:smooth-table1}, we show a brief description of each model (second column, while the DeepONet architectures are shown in Table \ref{table:architectures}). The total number of trainable parameters for all models is shown in the third column and the relative $L_2$ error on the test dataset in the fourth column. In addition, we show the relative error for the two out-of-distribution data sets (OOD$_1$ and OOD$_2$ from Table \ref{table:datasets}). Finally, we evaluate the model performance on a noisy dataset which consists of the original test dataset with $10\%$ added uncorrelated Gaussian noise (last column). 

\begin{table}[ht!]
\scriptsize
\caption{Relative $L_2$ error and number of trainable model parameters for Case I of the Brusselator model for $N=16,000$ training data and $N_* = 4,000$ testing data.}
\centering
\begin{tabular}{c c c c c c c}
\toprule
\multirow{2}{*}{Method} & \multirow{2}{*}{Description} & \multirow{2}{*}{$\#$ of params} & \multicolumn{4}{c}{Relative $L_2$ error} \\ \cmidrule(l){4-7} 
& & & Test data  & OOD$_1$ & OOD$_2$ & $+10\%$ noise data \\
 \toprule
U.\ m-PCE & $d_{\text{in}}=18$, $d_{\text{out}}=20$ & $3,800$ & $5.28 \pm 0.02 \%$ & $5.42 \pm 0.01 \%$ & $4.65 \pm 0.04 \%$  & $5.30 \pm 0.04 \%$ \\
O.\ m-PCE & $d_{\text{in}}=30$, $d_{\text{out}}=45$ & $22,320$ & $4.03 \pm 0.02 \%$ & $4.18 \pm 0.02 \%$ & $3.60 \pm 0.03 \%$  &  $4.15 \pm 0.03 \%$ \\ \hdashline
DeepONet & (see Table \ref{table:architectures}) & $127,740$ &  $\mathbf{2.85 \pm 0.07 \%}$ & $3.07 \pm 0.07 \%$ & $2.92 \pm 0.07 \%$ & $2.90 \pm 0.09 \%$ \\  \hdashline
FNO & modes=[8,8,11] & 46,158,529 & $3.08 \pm 0.08 \%$ & $4.57 \pm 0.11 \%$ & $4.01 \pm 0.09 \%$ & $17.24 \pm 0.08 \%$  \vspace{1pt} \\ 
\bottomrule
\end{tabular}
\label{table:smooth-table1}
\end{table}

\begin{figure}[ht!]
\begin{center}
\includegraphics[width=0.9\textwidth]{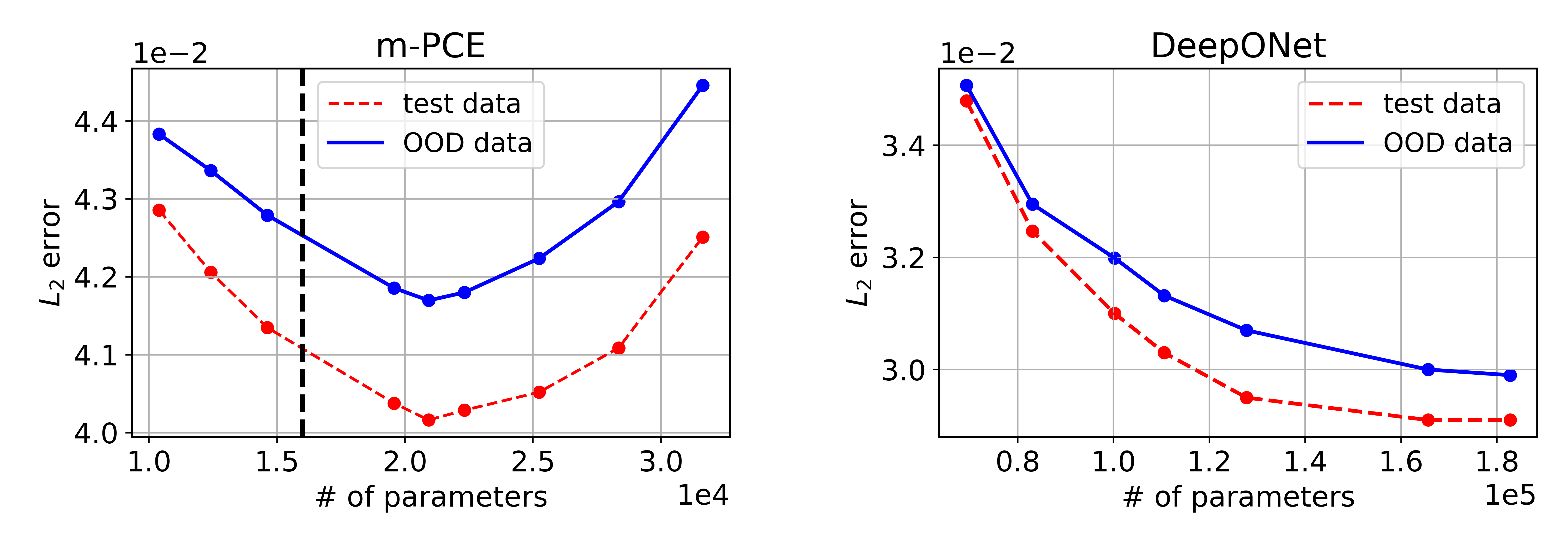}
\caption{Relative $L_2$ error of m-PCE (left) and DeepONet (right) for interpolation (test data) and extrapolation (OOD$_1$ data) as a function of the total number of trainable parameters, for $N=16,000$ (Case I). A sweet spot between generalization and memorization is observed for $22,320$ trainable parameters for m-PCE. The vertical black line (left plot) represents the over/under-parameterization threshold, which is equal to the number of training data $N$. The DeepONet, on the other hand, shows continuous improvement with increasing number of parameters even over a much wider range of parameter sizes.}
\label{fig:m-PCE-input-dim}
\end{center}
\end{figure}
For the under-parameterized m-PCE surrogate, a relatively low dimensional embedding of the input, $d_{\text{in}}$ and output data, $d_{\text{out}}$ has been chosen to reduce the number of trainable parameters, which results in a considerably higher $L_2$ relative error compared to its over-parameterized counterpart. k-PCA\footnote{The python package \textit{scikit-learn} \cite{scikit-learn} was used for the implementation of k-PCA.} has been used as the DR method, with a radial basis function (`rbf') and a polynomial (`poly') kernel for the dimensionality reduction of the input and output data, respectively. In both under- and over-parameterized models, the embedding dimension has been optimized manually using a grid search. Among all models, the DeepONet exhibits the best performance with the smallest relative error $2.95 \pm 0.07 \%$. However, we note that all models perform comparably and reasonably well on the test and OOD data, pointing to their generalizability for this case. We further observe that all models perform better on OOD$_2$, which has a larger lengthscale than the training data set, than they do for OOD$_1$. This follows intuitively because a shorter lengthscale for the input random field increases its intrinsic dimension and makes the mapping more complex. When $10\%$ added noise is introduced to the training data, the m-PCE and DeepONet results are consistent with their noise-free errors, while the FNO error increases substantially. We show the performance of both DeepONet and m-PCE for higher levels of noise ($20\%$ and $30\%$) in Table \ref{table:smooth-table1_effect_regularizer}. Moreover, to examine the stability of the non-linear mappings we learnt, we added noise to the inputs of the testing data and we show these results in Table \ref{table:mPCE-noise}. The results in both tables suggest that there is no amplification of inference error even for high levels of input noise. 

To investigate further the influence of over/under parameterization in the considered model, we systematically study the m-PCE and DeepONet for increasing number of parameters. In Figure \ref{fig:m-PCE-input-dim}, we show the relative $L_2$ error for interpolation (test data) and extrapolation (OOD$_1$ data) for the over-parameterized m-PCE and DeepONet as a function of the number of trainable model parameters. We observe that for m-PCE (left plot) the error is minimized for $\sim 22,320$ trainable parameters and subsequently increases as the number of parameters increases. Thus, in m-PCE identifying the optimal number of parameters is crucial. For the DeepONet (right plot), we observe that the testing error continuously decreases as the number of model parameters increases, which implies that simply increasing network size improves performance, albeit with diminishing return.  


A closer look at the DeepONet architecture affirms the observation from Figure \ref{fig:m-PCE-input-dim}, with some interesting caveats. In Table \ref{table:smooth-table1_effect_regularizer}, the model parameters in the DeepONet are modified by altering the width of the FNN in the trunk net and also the depth of the fully connected layers beyond the convolutional layers in the branch net. Overall, we see moving down the table that as network complexity (number of parameters) decreases, the error increases. However, for very large networks having convolutional layers with sizes $[512,150]$, the error increases over the less complex networks with $[256, 150]$ convolutional layers. However, using a $L_2$ regularizer can improve the accuracy of the network. When a regularization with $\lambda= 0.001$ is used, the accuracy rather slightly deteriorated. However, when $\lambda =0.0008$, the accuracy of the network improves. This shows that with enormous over-parameterization, DNNs might require some appropriate regularization to avoid slight over-fitting. 

\begin{table}[ht!]
\scriptsize
\caption{Effect of network complexity and $L_2$ regularization in DeepONet: Relative $L_2$ error of DeepONet for different network architectures of the branch net and the trunk net for Case I of the Brusselator model for $N=16,000$ training data and $N_* = 4,000$ testing data. The convolution kernels and the filters in the CNN architecture are kept constant for all the simulations. The width of the FNN layers beyond the convolutional layers is altered along with the width of the trunk net while the depth of the network is constant. The noise is added in the input of the \emph{training} dataset.}
\centering
\begin{tabular}{c c c c c c c c}
\toprule
\multirow{2}{*}{Branch net} & \multirow{2}{*}{Trunk net} & \multirow{2}{*}{$\#$ of params} & $\lambda$ in L$2$ & \multicolumn{4}{c}{Relative $L_2$ error ($\%$)} \\ \cmidrule(l){5-8} 
& & &$regularizer$ & Test data  & $+10\%$ noise & $+20\%$ noise & $+30\%$ noise\\
 \toprule
$[512, 150]$ & $[128, 128, 128, 150]$ & $213353$ & \text{--} & $2.969$ & $2.972$  & $2.981$ & $3.00$ \\
& & & $0.001$ & $3.0313$ & $3.0315$  & $3.023$ & $3.036$ \\
& & & $0.0008$ & $\textbf{2.69}$ & $2.695$  & $2.7065$ & $2.722$ \\
$[512, 150]$ & $[100, 100, 100, 150]$ & $196220$ & \text{--} & $3.058$ & $3.06$  & $3.069$ & $3.085$ \\
& & & $0.001$ & $3.062$ & $3.063$  & $3.070$ & $3.083$ \\
& & & $0.0008$ & $2.71$ & $2.712$  & $2.72$ & $2.728$ \\
$[256, 150]$ & $[128, 128, 128, 150]$ & $158316$ & \text{--} & $2.752$ & $2.754$  & $2.765$ & $2.783$ \\
$[256, 150]$ & $[100, 100, 100, 150]$ & $141180$ & \text{--} & $2.918$ & $2.921$  & $2.93$ & $2.947$ \\
$[128, 150]$ & $[128, 128, 128, 150]$ & $130796$ & \text{--} & $2.952$ & $2.955$  & $2.964$ & $2.979$ \\
$[128, 150]$ & $[100, 100, 100, 150]$ & $113660$ & \text{--} & $3.079$ & $3.082$  & $3.09$ & $3.105$ \\
$[64, 150]$ & $[100, 100, 100, 150]$ & $99900$ & \text{--} & $3.246$ & $3.248$  & $3.527$ & $3.269$ \\ 
$[64, 150]$ & $[128, 128, 128, 150]$ & $117036$ & \text{--} & $3.193$ & $3.196$  & $3.203$ & $3.219$ \vspace{1pt} \\ 
\bottomrule
\end{tabular}
\label{table:smooth-table1_effect_regularizer}
\end{table}

\begin{table}[ht!]
\scriptsize
\caption{Relative $L_2$ error of surrogates on noisy data for Case I $\&$ Case II and for $N = 800\times n_t$ training data. The noise is added in the inputs of the \emph{testing} dataset.}
\centering
\begin{tabular}{c c c c c c c}
\toprule
& & \multirow{2}{*}{Model description}  & \multicolumn{4}{c}{Relative $L_2$ error ($\%$)} \\ \cmidrule(l){4-7} 
& & & Test data  & $+10\%$ noise & $+20\%$ noise & $+30\%$ noise \\
 \toprule
\multirow{2}{*}{Case I} & U.\ m-PCE  & $d_{\text{in}}=18$, $d_{\text{out}}=20$ & $5.28$ & $5.30$ & $5.38$  & $5.55$ \\
& O.\ m-PCE  & $d_{\text{in}}=30$, $d_{\text{out}}=45$ & $4.03$ & $4.15$ & $4.30$  &  $4.63$ \\ \hdashline 
\multirow{5}{*}{Case II} & U.\ m-PCE  & $d_{\text{in}}=25$, $d_{\text{out}}=40$ &  $7.69$ & $7.97$ & $8.92$ & $11.44$ \\  
& O.\ m-PCE  & $d_{\text{in}}=23$, $d_{\text{out}}=105$ & $6.04$ & $6.26 $ & $7.63$ & $8.50$ \\  
 & DeepONet & (see Table \ref{table:architectures}) & $4.22$ & $4.29$ & $4.52$ & $4.79$\\
& POD-DeepONet & (see Table \ref{table:architectures}) & $3.23$ & $3.26$ & $3.30$ & $3.37$\\
& SA-DeepONet & (see Table \ref{table:architectures}) & $3.18 $ & $3.20 $ & $3.26$ & $3.34$ \vspace{1pt} \\
\bottomrule
\end{tabular}
\label{table:mPCE-noise}
\end{table}

\begin{table}[ht!]
\footnotesize
\caption{Relative $L_2$ error for Case I of the Brusselator model showing the effect of dataset size.}
\centering
\begin{tabular}{c c c c c}
\toprule
\multirow{2}{*}{Training data} & \multicolumn{4}{c}{Relative $L_2$ error} \\ \cmidrule(l){2-5} & Under.\ m-PCE & Over.\ m-PCE & DeepONet & FNO \\
 \toprule
$100\times n_t$  & $5.75 \pm 0.02 \%$ & $5.35 \pm 0.04 \%$ & $6.03 \pm 0.10 \%$ & $11.07 \pm 0.23 \%$ \\
$200\times n_t$  &   $5.68 \pm 0.03 \%$ & $5.17 \pm 0.02 \%$& $5.47 \pm 0.09 \%$ & $6.16 \pm 0.27 \%$ \\
$400\times n_t$  &  $5.39 \pm 0.03 \%$ &  $4.62 \pm 0.03 \%$ & $ 3.72 \pm 0.11 \%$ & $4.13 \pm 0.13 \%$ \\  
$800\times n_t$ & $5.28 \pm 0.02 \%$ &  $4.03 \pm 0.02 \%$ & $2.85 \pm 0.07 \%$ & $3.08 \pm 0.08$ \\  $1600\times n_t$ & $5.22 \pm 0.04 \%$ & $3.69 \pm 0.01 \%$ & $\mathbf{2.41 \pm 0.08 \%}$ & $2.98 \pm 0.07 \%$ \vspace{1pt} \\
\bottomrule
\end{tabular}
\label{table:smooth-table2}
\end{table}

\begin{figure}[ht!]
\begin{center}
\includegraphics[width=1\textwidth]{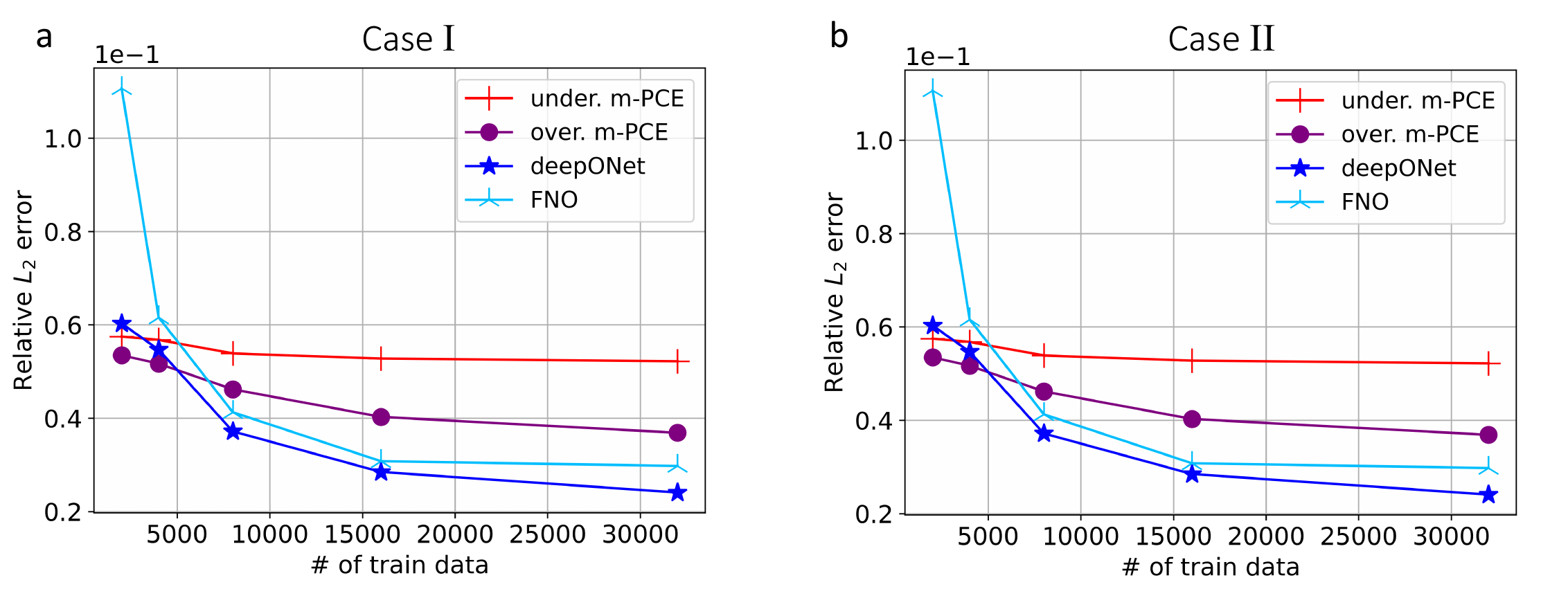}
\caption{Effect of dataset size for Case I (a) and Case II (b) of the Brusselator model. The relative $L_2$ error is shown for all models for five cases: $N=100\times n_t, N=200\times n_t, N=400\times n_t, N=800\times n_t, N=1600\times n_t$. Mean values with corresponding uncertainties are presented in Table \ref{table:smooth-table2} (Case I) and Table \ref{table:sharp-table2} (Case II).}
\label{fig:dataset-size}
\end{center}
\end{figure}

\begin{figure}[ht!]
\begin{center}
\includegraphics[width=1\textwidth, trim=0cm 0cm 0cm 3.91cm, clip]{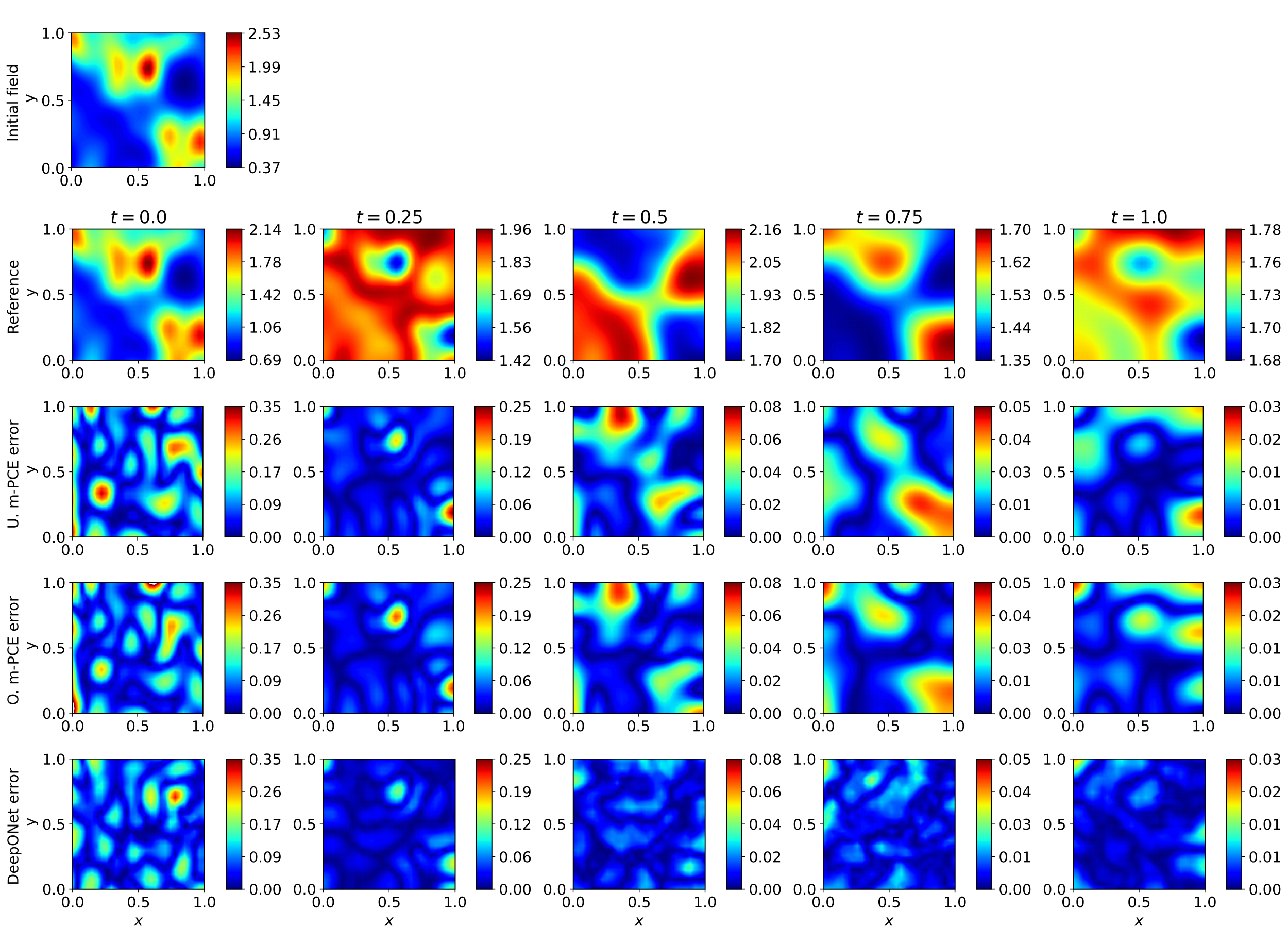}
\caption{(Case I) Reference model response (first row), under- and over-parameterized m-PCE prediction error (second and third row) and DeepONet prediction error (fourth row) for 5 snapshots, $\it{i.e.}$, $t=\{0, 0.25, 0.50, 0.75, 1\}$.}
\label{fig:comparison-smooth-m-PCE}
\end{center}
\end{figure}

Another way to look at over/under-parameterization of the models is by increasing the training data set size. As we increase the training data set size, the models become closer to being under-parameterized. In Table \ref{table:smooth-table2}, the relative $L_2$ error is shown for five different training data set sizes. 
For better visualization, we also plot the relative $L_2$ error in Figure \ref{fig:dataset-size}(a). We observe that, for all models, the error decreases as the training set size increases. However, the over-parameterized models (m-PCE, DeepONet, and FNO) have in general a comparable performance, while the under-parameterized m-PCE sees only minimal improvement. We further find that in cases where data are sparse ($N<250\times n_t$), m-PCE slightly outperforms the other models. However, for $N>250\times n_t$, DeepONet exhibits the best performance, with the lowest relative error $2.41 \pm 0.08 \%$ for $N=1600\times n_t$ data, with the FNO showing similar accuracy. We further recognize that the over-parameterized m-PCE, in fact, becomes under-parameterized after $N$ grows beyond 22,320. The DeepONet and FNO, on the other hand, are still vastly over-parameterized even in the largest data set cases. These results show that the over-parameterized models have inbuilt flexibility to continue learning as data set size increases, while the under-parameterized models reach their training limit. However, the under-parameterized models are superior for small training datasets.

For illustration purposes, the relative errors for the U m-PCE, O m-PCE, and DeepONet are shown for 5 time snapshots $t=\{0,0.25,0.5,0.75,1\}$ from a given initial condition in Figure \ref{fig:comparison-smooth-m-PCE}. Despite the high variability in the input fields and the complexity of the model solution, m-PCE and DeepONet are able to accurately predict the model response for all time steps with small local errors that are observed in areas where there is a rapid change in the model response.

Finally, in terms of computational cost, training the m-PCE is significantly faster. For $N=800\times n_t$ data the cost of training the m-PCE (including both the manifold learning and surrogate construction) was $1.85s$ on a $2.8$GHz quad-core Intel i7 CPU. Meanwhile, the DeepONet took $3.42h$ and the FNO $5.26h$ for training, both trained on an NVIDIA A6000 GPU. For the DeepONet and FNO, the batch size is 200.  We therefore see that the m-PCE training is $\sim7,000$ and $\sim10,000$ times faster than the DeepONet and FNO, respectively. Nonetheless, after the training procedure, all models are capable of making predictions in a fraction of a second.

\subsection{Case II}
\label{CaseII}

Similar to Section \ref{CaseI}, we test the model performance for $N=800\times n_t=8,000$ training data for Case II. In this case, the system is enters a limit cycle and thus after sufficient time we observe periodic oscillations. The sharp drop in the model response, as shown in Section \ref{data-generation}, is particularly challenging for the surrogates to learn. To address this challenge, we additionally employ POD-DeepONet, and the proposed SA-DeepONet. The predictive accuracy of all models are shown in Table \ref{table:sharp-table1}. We observe that for Case II, the difference between m-PCE and the DeepONets is more significant than in Case I. SA-DeepONet, in particular, achieves a very small relative error $L_2 = 3.19 \pm 0.02 \%$ and shows very good performance for both in- and out-of-distribution data, and also when Gaussian noise is added to the input random fields. Consistent with the results of the previous section, the over-parameterized m-PCE achieves significantly higher predictive accuracy than its under-parameterized counterpart, and the error does not significantly increase when tested on OOD and noisy data. On the other hand, the prediction accuracy of FNO is comparable for out-of-distribution datasets, however, for noisy inputs ($+10\%$ noise), the network fails to approximate reasonably, with a very high relative $L_2$ error of $29.61 \pm 0.3\%$. The performance of all surrogates models for higher levels of noise ($+20\%$ and $+30\%$) is shown in Table \ref{table:mPCE-noise}. Compared to DeepONet models, mPCE shows more significant deterioration in performance, however, within acceptable limits.

\begin{table}[ht!]
\footnotesize
\caption{Relative $L_2$ error for Case II of the Brusselator model for $N=800\times n_t$ training data.}
\centering
\begin{tabular}{c c c c c c}
\toprule
\multirow{2}{*}{Method} & \multirow{2}{*}{Description} & \multicolumn{4}{c}{Relative $L_2$ error} \\ \cmidrule(l){3-6} &  & Test data  & OOD data$_1$ & OOD data$_2$ & $+10\%$ noise data \\
 \toprule
Under.\ m-PCE & $d_{\text{in}}=25$, $d_{\text{out}}=40$ & $7.69 \pm 0.02 \%$ & $8.39 \pm 0.01 \%$ & $8.19 \pm 0.04 \%$ & $7.97 \pm 0.04 \%$ \\
Over.\ m-PCE & $d_{\text{in}}=23$, $d_{\text{out}}=105$ & $6.04 \pm 0.03 \%$ & $6.88 \pm 0.02 \%$ & $6.47 \pm 0.03 \%$ & $6.26 \pm 0.04 \%$ \\ \hdashline
DeepONet & (see Table \ref{table:architectures}) &  $4.09 \pm 0.10 \%$ & $5.35 \pm 0.09 \%$ & $4.57 \pm 0.08 \%$  & $4.10 \pm 0.11 \%$ \\  
POD-DeepONet & (see Table \ref{table:architectures}) &  $3.36 \pm 0.01 \%$ & $4.50 \pm 0.05 \%$ & $3.36 \pm 0.06 \%$  & $3.37 \pm 0.06 \%$ \\
SA-DeepONet & (see Table \ref{table:architectures}) & $\mathbf{3.19 \pm 0.02 \%}$ & $4.76 \pm 0.02 \%$ & $3.55 \pm 0.05 \%$  & $3.20 \pm 0.03 \%$ \\ \hdashline
FNO & modes=[8,8,6] & $4.17 \pm 0.21 \%$ & $6.18 \pm 0.55 \%$ & $5.92 \pm 0.31 \%$ & $29.61 \pm 0.30\%$ \vspace{1pt} \\ 
\bottomrule
\end{tabular}
\label{table:sharp-table1}
\end{table}

\begin{table}[ht!]
\scriptsize
\caption{Relative $L_2$ error for Case II of the Brusselator model showing the effect of dataset size.}
\centering
\begin{tabular}{c c c c c c c}
\toprule
\# of data & \multicolumn{6}{c}{Relative $L_2$ error} \\ \cmidrule(l){2-7} & U.\ \ m-PCE & O.\ \ m-PCE & DeepONet & POD-DeepONet & SA-DeepONet & FNO \\
 \toprule
$100\times n_t$  & $10.46 \pm 0.03 \%$ & $8.66 \pm 0.04 \%$ & $8.83 \pm 0.08 \%$ & $8.14 \pm 0.02 \%$ & $7.78 \pm 0.02 \%$ & $13.64 \pm 0.87 \%$ \\
$200\times n_t$  &  $8.41 \pm 0.06 \%$ & $7.55 \pm 0.03 \%$& $7.21 \pm 0.10 \%$ & $5.56 \pm 0.01 \%$ & $6.06 \pm 0.03 \%$ & $8.77 \pm 0.18 \%$ \\
$400\times n_t$  &  $7.91 \pm 0.06 \%$ &  $7.35 \pm 0.01 \%$ & $ 5.38 \pm 0.09 \%$ & $5.03 \pm 0.02 \%$ & $3.86 \pm 0.04 \%$ & $6.25 \pm 0.42 \%$ \\  
$800\times n_t$ & $7.69 \pm 0.02 \%$ &  $6.04 \pm 0.03 \%$ & $4.09 \pm 0.10 \%$ & $3.36 \pm 0.01 \%$ & $3.19 \pm 0.02 \%$ & $4.17 \pm 0.21 \%$ \\  
$1600\times n_t$ & $6.77 \pm 0.04 \%$ & $5.41 \pm 0.06 \%$ & $3.78 \pm 0.09 \%$ & $3.28 \pm 0.03 \%$ & $\mathbf{2.93 \pm 0.03 \%}$ & $3.78 \pm 0.28 \%$ \vspace{1pt} \\
\bottomrule
\end{tabular}
\label{table:sharp-table2}
\end{table}

The relative $L_2$ errors for Case II follow the same observations as made in Case I. The surrogate model approximates the field more accurately for OOD$_2$ over OOD$_1$. The intrinsically non-smooth nature of the solution in this cases poses a fundamental problem in developing a surrogate model. Most surrogate models, such as m-PCE, typically demand smoothness in the system response, and therefore we observe higher relative $L_2$ errors in both the under and over-parameterized m-PCE. However, using the DeepONet, the locations of sharp features can be manually selected and penalized along with the regular loss function, which improves optimization of the model. The SA-DeepONet, meanwhile, makes this task easier by automatically selecting and updating the penalty parameters, to give the most accurate solution. 

As displayed in Table \ref{table:sharp-table2} and also shown in Figure \ref{fig:dataset-size}(b), we once again see that the dataset size affects significantly the results for all models. Similar to Case I, the error continuously decreases with increasing training set size for all models. However, in this case due to the non-smooth solution, the DeepONet provides superior performance even for small data sets and the SA-DeepONet shows the best overall performance. We further see that the DeepONet models converge very rapidly and do not show significant improvement for much larger data sets.

\begin{figure}[ht!]
\begin{center}
\includegraphics[width=1\textwidth, trim=0cm 0cm 0cm 3.91cm, clip]{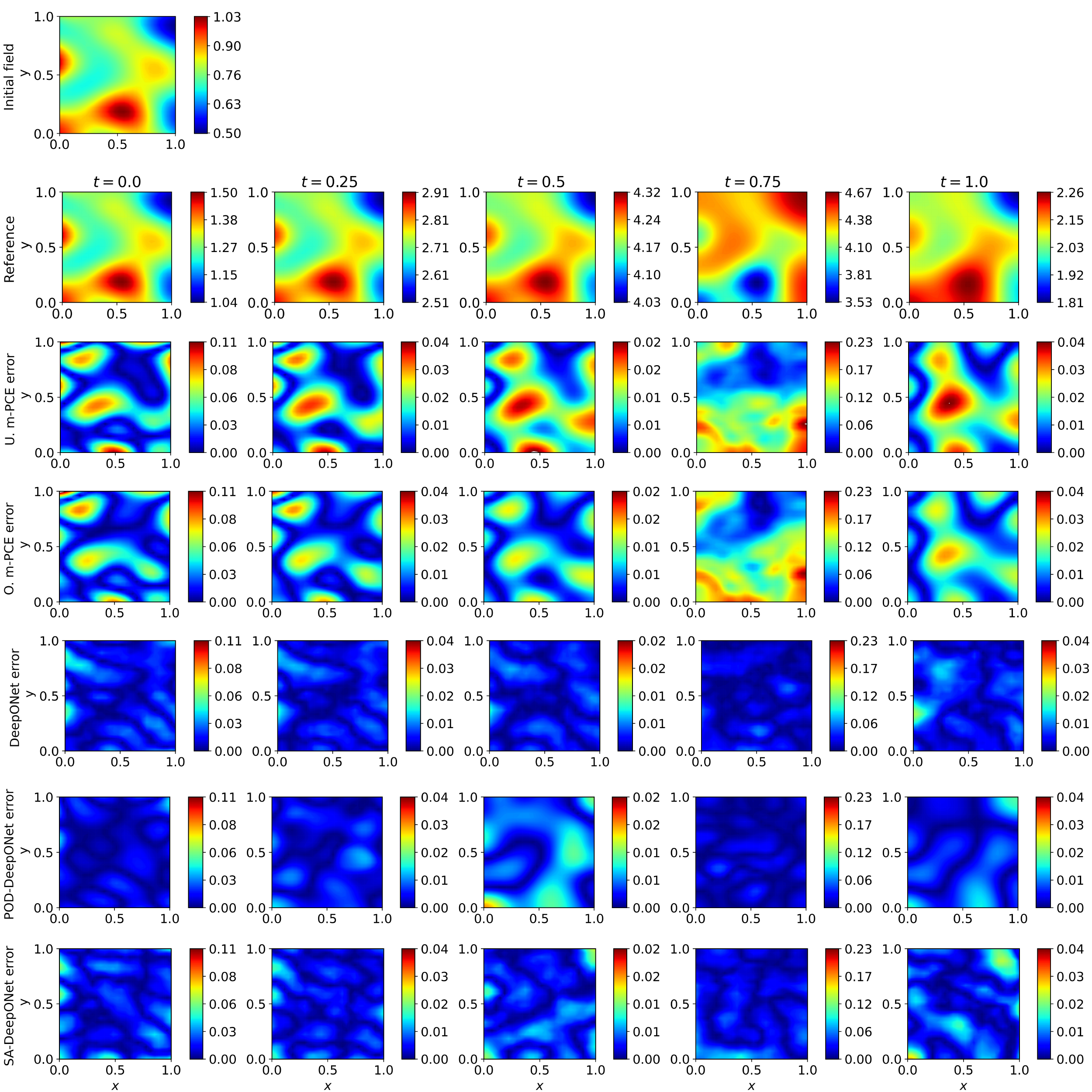}
\caption{(Case II) Reference model response (first row), under- and over-parameterized m-PCE prediction error (second and third fourth rows) and DeepONet, POD-DeepONet and SA-DeepONet prediction error (fourth, fifth, and sixth rows) for 5 snapshots, $\it{i.e.}$, $t=\{0, 0.25, 0.50, 0.75, 1\}$.}
\label{fig:comparison-sharp-m-PCE}
\end{center}
\end{figure}

Once again for illustration, we show representative relative error plots for m-PCE, standard DeepONet, POD-DeepONet and SA-DeepONet for a given random realization of the input stochastic field in Figure \ref{fig:comparison-sharp-m-PCE}. Both the under- and over-parameterized m-PCE provide a good approximation of the evolving field, however, when the sharp transition caused by temporal oscillations occurs ($t=0.75$), the error grows substantially. The DeepONet models show a greater overall consistency between the prediction and ground truth, while they also handle more effectively the non-smoothness in the model response, leading to a smaller overall test error.

Although m-PCE performed satisfactorily in both dataset cases, we note that there are many possibilities for improvement of the method which were not explored here. Only the k-PCA method was employed in this study for lower-dimensional embedding of both input and output quantities, but the flexibility of m-PCE allows any suitable manifold learning technique to be employed. The choice of which method to use should be based on the complexity and nonlinearity of the model. Exploring other manifold learning methods may, in fact, improve the error for this problem significantly. In addition, improvements in the PCE model construction, e.g., via adaptive refinement strategies such as Least Angle Regression \cite{blatman2011adaptive} or through adaptive refinement with multi-element PCE \cite{wan2006multi} could improve its treatment of non-smooth models. These warrant further investigation and are increasingly motivated by the fact that the m-PCE is significantly less computationally expensive, with training time on the order of seconds on a CPU versus minutes or hours for the DeepONet.

\section{Summary}
\label{sec:summary_and_discussions}

In this study, we have investigated and compared the performance of under-and over-parameterized manifold-based surrogates with over-parameterized deep neural operators, specifically DeepONet and FNO, for approximating complex, nonlinear, and high-dimensional models. To compare the surrogate models, we have considered the Brusselator reaction-diffusion system, which is modeled by a 2D time-dependent PDE, where the initial field of one of the species is modeled as a Gaussian random field. The surrogate models at hand are investigated in terms of relative error for a relatively smooth and a highly nonsmooth dataset, and tested on three regression tasks: interpolation, extrapolation, and robustness to noise. Our observations can be summarized as follows:
\begin{enumerate}
\item Over-parameterization of models lead to the construction of more expressive surrogates, achieving higher (often significantly) generalization accuracy than their under-parameterized counterparts (See Tables \ref{table:smooth-table1} and \ref{table:sharp-table1} fourth column). DeepONet is less sensitive to the total number of trainable parameters, as the extrapolation performance is not significantly affected when this number is varied. On the other hand, m-PCE is more sensitive to the total number of trainable parameters, whether this number is controlled by the input or output embedding dimensionality or the maximum polynomial degree. In cases where the Brusselator model exhibits a relatively smooth response (Case I), we found that the over-parameterized surrogate models in general show comparable performance, while achieving a very small generalization error for both in- and out-of-distribution inputs as well as noisy inputs (See Table \ref{table:smooth-table1}). 
\item When highly non-smooth dynamics are considered, DeepONet and its extensions (POD-DeepONet and weight self-adaptivity DeepONet) exhibited very high generalization accuracy. While all three DeepONet variants (standard, POD, and SA) performed very well in Case II in both interpolation and extrapolation, the SA-DeepONet outperformed the rest and is found to be very robust to noise. (See Table \ref{table:sharp-table1}). On the other hand, m-PCE performed well overall but failed to approximate accurately the system response near the sharp transition. This challenge could be addressed by employing enhanced PCE approaches such as multi-element PCE \cite{wan2005adaptive}. 
\item In general, all surrogate models (under- and over-parameterized) generalize the solution for OOD data having length scales higher than the training data more accurately than for OOD data having lower length scales, which results in higher variability in the input fields and thus model response (See tables \ref{table:smooth-table1} and \ref{table:sharp-table1} fifth and sixth columns).
\item The robustness of the surrogate models is evaluated by testing the predictions with noisy inputs. Additionally, we observe the consistency of the surrogate models to make appropriate approximations under the circumstance of noisy input data. We interpolated smoothly between the cases of no noise and inducing $10\%$, $20\%$ and $30\%$ noise to observe a very small deterioration of the generalization error as the noise level is increased. We found that both DeepONet and m-PCE  surrogates can capture the remaining signal in the data while also fitting the noisy input data (see Tables \ref{table:smooth-table1}, \ref{table:smooth-table1_effect_regularizer}, \ref{table:smooth-table2},\ref{table:mPCE-noise}), while FNO failed to perform well in such cases.
\item The effect of dataset size has significant impact on the accuracy of the over-parameterized surrogate models. For smooth response (Case I), the under-parameterized model is not significantly improved by the increase in dataset size, while for all the over-parameterized models, the accuracy improves as the training dataset size increases. DeepONet exhibits the highest level of accuracy for large datasets, however, when small datasets are considered, over-parameterized m-PCE outperforms DeepONet (see Figure \ref{fig:dataset-size}). Considering non-smooth dynamics (Case II), under-parameterized models fail to capture the response accurately while the over-parameterized models exhibit similar behaviour as discussed for Case I (see Table \ref{table:sharp-table2}).
\item DeepONet and its proposed extensions exhibit the highest generalization accuracy overall for varying degrees of complexity in model response and provide an expressive surrogate modeling method with the ability to learn the nonlinear operator, thus allowing for predicting model response for new unseen input functions, boundary conditions, domain geometries etc, at high training cost. 
\item The m-PCE provides a powerful and flexible surrogate modeling approach, which combines high predictive accuracy and robustness to noise with an extremely low training computational cost compared to all other methods. This makes it particularly attractive for tasks involving active learning methods that iteratively retrain the surrogate. Finally, significant research potential exists to improve m-PCE methods to achieve better generalization for cases of non-smooth, high-dimensional models.
\end{enumerate}

\section*{Acknowledgements}
For KK and MDS, this material is based upon work supported by the U.S. Department of Energy, Office of Science, Office of Advanced Scientific Computing Research under Award Number DE-SC0020428.
SG and GEK would like to acknowledge support by the DOE project PhILMs (Award Number DE-SC0019453) and the OSD/AFOSR MURI grant FA9550-20-1-0358.
\bibliographystyle{elsarticle-num} 
\bibliography{cas-refs}

\appendix
\section{Model Architectures}

\begin{table}[ht!]
\small
\caption{DeepONet architectures for the Brusselator problem Case I and Case II. In the third column, $d$ represents the depth of the network, $w$ the width, $p$ the latent dimension at the last layer and $m$ the number of POD modes.}
\centering
\begin{tabular}{c c c c}
\toprule
 Model & Branch net & \begin{tabular}{@{}c@{}}Trunk net/ \\ no.\ of POD-modes\end{tabular} & \begin{tabular}{@{}c@{}}Activation \\ function\end{tabular}  \\
 \toprule
DeepONet (Case I)  & CNN & $d=4 , w=128, p=150$ & ReLU  \\
DeepONet (Case II)  &  CNN & $d=4 , w=128, p=150$ & ReLU  \\
POD-DeepONet (Case II)  &  CNN &  modes $=100, p=100$ & ReLU \\  
SA-DeepONet (Case II) & $d=3 , w=128, p=100$ &  $d=4 , w=128, p=100$ & ReLU    \\
\bottomrule
\end{tabular}
\label{table:architectures}
\end{table}

\end{document}